\documentclass[twoside,11pt]{article}
\usepackage{jair, theapa, rawfonts}
\usepackage{amsthm}
\usepackage{tabularx}
\usepackage{booktabs}
\usepackage{floatrow}
\usepackage{float}
\usepackage{subfig}
\usepackage{amsmath, amssymb}
\usepackage{bm}
\usepackage{array,multirow,graphicx}
\usepackage{multirow}
\usepackage{eufrak}
\usepackage{boldline}
\usepackage[linesnumbered,ruled,vlined]{algorithm2e}
\usepackage[table, xcdraw]{xcolor}
\usepackage{fontawesome}
\usepackage{colortbl}
\usepackage{wrapfig}
\usepackage{url}
\usepackage{breakurl}
\usepackage{footnote}
\makesavenoteenv{tabular}
\usepackage{amsmath}
\usepackage{amsfonts}
\usepackage{amssymb}

\usepackage[switch]{lineno}

\newcommand\doubleRule{\toprule\toprule}

\usepackage{algorithmicx}
\usepackage{multirow}
\usepackage{multicol}
\usepackage{booktabs}
\usepackage{array}
\usepackage{xspace}
\usepackage{bigstrut}

\usepackage{appendix}

\usepackage{stfloats}
\usepackage[T1]{fontenc}
\usepackage{xcolor}
\usepackage{wrapfig}
\SetKwInput{KwInput}{Input}
\SetKwInput{KwOutput}{Output} 
\SetKwInput{Kwall}{all} 
\newtheorem{theorem}{Theorem}
\newtheorem{lemma}{Lemma}

\ShortHeadings{FlexiBO: Decoupled Cost-Aware Multi Objective Optimization}
{Iqbal, Su, Kotthoff, \& Jamshidi}
\firstpageno{1}
\usepackage{flafter}
\usepackage{todonotes}

\newcommand{\tool}{\textsc{FlexiBO}\xspace}
\newcommand{\pal}{\textsc{PAL}}
\newcommand{\pesmo}{\textsc{PESMO}}
\newcommand{\camobo}{\textsc{CA-MOBO}}
\newcommand{\parego}{\textsc{ParEGO}}
\newcommand{\smsego}{\textsc{SMSego}}
\newcommand{\mesmo}{\textsc{MESMO}}
\newcommand{\mesmoc}{\textsc{MESMOC}}
\newcommand{\pesmodec}{\textsc{PESMO-DEC}}
\newcommand{\epal}{$\epsilon$-\textsc{PAL}}
\newcommand{\gplc}{\textsc{FlexiBO-GPLC}}
\newcommand{\gprc}{\textsc{FlexiBO-GPRC}}
\newcommand{\gpcc}{\textsc{FlexiBO-GPCC}\xspace}

\newcommand{\rflc}{\textsc{FlexiBO-RFLC}}

\usepackage{subfig}
\usepackage{pifont}
\newcommand{\cmark}{\ding{51}}%
\newcommand{\xmark}{\ding{55}}%
\usepackage{enumitem}

\begin{document}

\title{\tool: A Decoupled Cost-Aware Multi-Objective Optimization Approach
       for Deep Neural Networks}

\author{\name Md Shahriar Iqbal \email miqbal@email.sc.edu \\
       \addr University of South Carolina, Columbia, SC, USA\\
       \AND
       \name Jianhai Su \email suj@email.sc.edu \\
       \addr University of South Carolina, Columbia, SC, USA\\
       \AND
       \name Lars Kotthoff \email larsko@uwyo.edu \\
       \addr University of Wyoming, Laramie, WY, USA\\
       \AND
       \name Pooyan Jamshidi \email pjamshid@cse.sc.edu \\
       \addr University of South Carolina, Columbia, SC, USA}


\maketitle
\begin{abstract}
The design of machine learning systems often requires trading off different objectives, for example, prediction error and energy consumption for deep neural networks (DNNs). Typically, no single design performs well in all objectives; therefore, finding Pareto-optimal designs is of interest. The search for Pareto-optimal designs involves evaluating designs in an iterative process, and the measurements are used to evaluate an acquisition function that guides the search process. However, measuring different objectives incurs different costs. For
example, the cost of measuring the prediction error of DNNs is orders of magnitude higher than that of measuring the energy consumption of a pre-trained DNN as it requires re-training the DNN. Current state-of-the-art methods do not consider this difference in objective evaluation cost, potentially
incurring expensive evaluations of objective functions in the optimization process. In this paper, we develop a novel decoupled and cost-aware multi-objective optimization algorithm, we call \emph{\underline{Flexi}ble Multi-Objective \underline{B}ayesian \underline{O}ptimization} (FlexiBO) to address this issue. For evaluating each design, FlexiBO selects the objective with higher relative gain by weighting the improvement of the hypervolume of the Pareto region with the
measurement cost of each objective. This strategy, therefore, balances the expense of collecting new information with the knowledge gained through objective evaluations, preventing FlexiBO from performing expensive measurements for little to no gain. We evaluate FlexiBO on seven state-of-the-art DNNs 
for image recognition, natural language processing (NLP),
and speech-to-text translation. Our results indicate that, given the same total experimental budget, FlexiBO discovers designs with 4.8$\%$ to 12.4$\%$ lower hypervolume error than the best method in state-of-the-art multi-objective optimization.  
\end{abstract}

\section{Introduction}
\begin{figure}[t]
\centering
  \subfloat[a][DNN System Stack]{\includegraphics[width=0.65\textwidth]{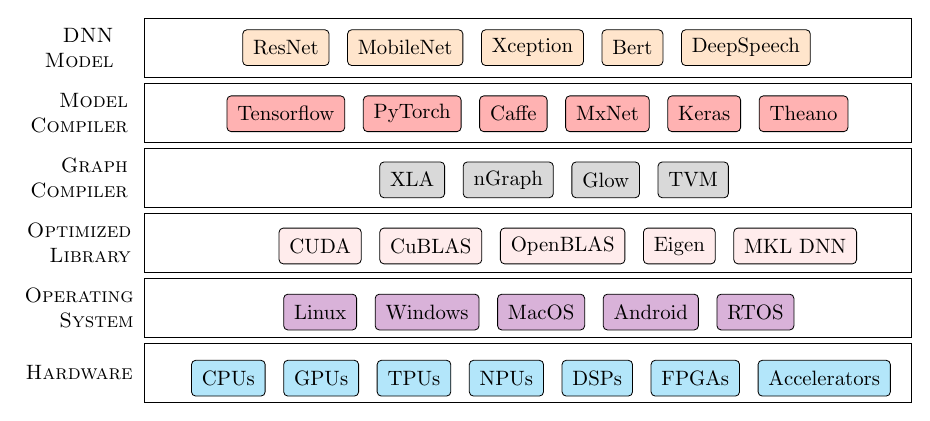} \label{fig:sys_stack}} 
  \centering
  \subfloat[b][Performance distribution with different design options]{\includegraphics[width=0.38\textwidth]{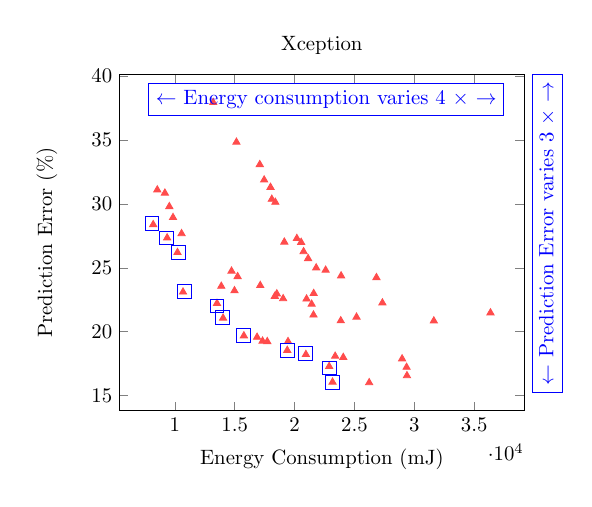} \label{fig:trade_offs}} 
 
  \caption{(a) The system stack of a DNN, composed of six components. (b) Performance variation of several hundred designs for an image recognition DNN system, \textsc{SqueezeNet}, deployed on resource-constrained \textsc{Nvidia Jetson TX2} hardware while performing inference on 5,000 test images.  We observe that there is substantial variation for each performance objective across different designs, i.e., prediction error varies 3$\times$, energy consumption varies 4$\times$. Designs marked with blue rectangles optimally trade off both objectives.}
\end{figure}

Recent developments of deep neural networks (DNNs) have sparked a growing demand for pushing the deployment of artificial intelligence applications from the cloud to a wide variety of edge and IoT devices. As these devices are closer to data and information generation sources, they provide better user experience e.g., latency and throughput sensitivity, security etc. Compared to datacenters, the edge devices are more resource-constrained and may not be even able to host these compute expensive DNN models. Therefore, designing energy-efficient DNNs is critical for successful deployment of DNNs to these devices with limited resources. On one hand, high inference error often leads to application failures~\shortcite{pei2017deepxplore,sun2018concolic}, but on the other hand, it is essential to reduce the number of computation cycles and/or memory footprints of DNNs to conserve energy~\shortcite{sze2017efficient}. In addition to DNN models, a number of components in the DNN system stack (shown in Figure~\ref{fig:sys_stack}) must work together for the seamless deployment of energy-efficient DNNs without compromising inference accuracy~\shortcite{gadepally2019ai,reuther2019survey}.

There exist 100s if not 1000s of design options from each component across the DNN system stack that impact the computational and memory requirements of DNNs and make them difficult to effectively deploy.  One of the key challenges in designing an optimal DNN system is to efficiently explore the vast design space, with non-trivial interactions of options from different components across the system stack, e.g., CPU frequency, GPU frequency, number of epochs, etc. Additionally, there is usually no single design that performs well for all performance objectives (see e.g.\ Figure~\ref{fig:trade_offs}). Therefore, we need to identify designs that provide optimal trade-offs -- \emph{Pareto optimal} designs. 

\begin{figure}[t!]
    \centering
    \resizebox{\linewidth}{!}{%
    \begin{tabular}{|lcccc|}
     \hline
     \hline 
     \multicolumn{1}{|l}{\textsc{Method}}& \rotatebox{0}{\textsc{Method Type}} & \rotatebox{0}{\textsc{Search Strategy}} & \rotatebox{0}{\textsc{Evaluation Strategy}} & \rotatebox{0}{\textsc{Cost Awareness}} \\ 
     \hline 
     
     \multicolumn{1}{l}{} & \multicolumn{1}{l}{} & \multicolumn{1}{l}{}&
     \multicolumn{1}{l}{} &\multicolumn{1}{l}{} \\[-0.9em] \hlineB{2.5}
     
     \multicolumn{1}{V{2.5}l}{\textsc{NEMO}}~\shortcite{kim2017nemo} & NAS&Gradient Based&Coupled&\xmark\\
     \multicolumn{1}{V{2.5}l}{\textsc{DPP-NET}}~\shortcite{dong2018dpp} & NAS&Gradient Based&Coupled&\xmark\\
     \multicolumn{1}{V{2.5}l}{\textsc{HR}}~\shortcite{liu2017hierarchical} & NAS&Gradient Based&Coupled&\xmark\\
     \multicolumn{1}{V{2.5}l}{\parego}~\shortcite{knowles2006parego} & MOBO&Random Scalarization&Coupled&\xmark\\
     \multicolumn{1}{V{2.5}l}{\smsego}~\shortcite{ponweiser2008multiobjective} & MOBO&Hypervoume Improvement&Coupled&\xmark\\
     \multicolumn{1}{V{2.5}l}{\pal}~\shortcite{zuluaga2013active} & MOBO&Predictive Uncertainty&Coupled&\xmark\\
     \multicolumn{1}{V{2.5}l}{\mesmo}~\shortcite{belakaria2019max} & MOBO&Output Space Entropy&Coupled&\xmark\\
     \multicolumn{1}{V{2.5}l}{\pesmo}~\shortcite{hernandez2015predictive} & MOBO&Input Space Entropy&Coupled&\xmark\\
     \multicolumn{1}{V{2.5}l}{\pesmodec}~\shortcite{hernandez2015predictive} & MOBO&Input Space Entropy&De-coupled&\xmark\\
     \multicolumn{1}{V{2.5}l}{\camobo}~\shortcite{abdolshah2019cost} & MOBO&Chebyshev Scalarization&Coupled&\cmark\\
     \hline
     \multicolumn{1}{V{2.5}l}{\tool}  & MOBO&Volume of the Pareto region&De-coupled&\cmark\\
    
     
     
      

    
    \hline
\end{tabular}}%

\caption{Comparison of \tool \ to related state-of-the-art methods that can be used to identify Pareto optimal solutions.}
\label{tab:related_work}
\end{figure}
Previous work has focused on neural architecture search (NAS) techniques that can efficiently locate Pareto optimal designs. NAS approaches like~\textsc{Nemo}~\shortcite{kim2017nemo}, hierarchical representations (HR)~\shortcite{liu2017hierarchical}, and \textsc{DPP-NET}~\shortcite{dong2018dpp} can be categorized according to three different criteria: (i) the \emph{Search Space}, (ii) the \emph{Optimization Method}, and (iii) the \emph{Candidate Evaluation Method}. Unfortunately, the effectiveness of NAS approaches largely depends on selecting a quality search space to reduce the complexity of search that requires significant prior knowledge that is difficult to find in practice, which also indicates that they are not suitable in different platforms. Much recent work has focused on multi-objective Bayesian optimization (MOBO) approaches like \textsc{ParEGO} \shortcite{knowles2006parego}, \textsc{SMSego} \shortcite{ponweiser2008multiobjective}, \textsc{PAL} \shortcite{zuluaga2013active}, \textsc{PESMO} \shortcite{hernandez2016predictive}, \textsc{MESMO} \shortcite{belakaria2019max}, \textsc{MESMOC} \shortcite{belakaria2020max} etc.\ to find the Pareto optimal designs that can used for hyperparameter tuning. MOBO approaches iteratively use the uncertainty captured by a probabilistic model (also known as the \emph{surrogate model}, an approximation that is much faster and cheaper to evaluate than the complex unknown function to be optimized) of the process to be optimized to compute the values of an \emph{acquisition function}. The optimum of the acquisition function provides an effective heuristic for identifying a promising design for which to evaluate the objectives. Nevertheless, there are limitations to these MOBO approaches. For instance, scalarization-based approaches (\parego) tend to suffer from sub-optimality, algorithms to optimize hypervolume based acquisition function (\smsego) scale poorly when the input dimensionality increases, methods that rely on entropy-based acquisition functions (\pesmo, \mesmo, \mesmoc) are computationally expensive, and \pal \ is simple to design but also requires a lot of computation at each iteration. 

Furthermore, most of these methods are cost unaware and consider the objective evaluation costs to be uniform. In practice, the cost of objective evaluations can be non-uniform e.g., optimizing prediction error in DNN systems is much more expensive than making predictions with a pre-trained DNN with different deployment system design options as that involves re-training the whole DNN.  Besides, existing approaches use a \emph {coupled evaluation strategy} to evaluate the design selected by the optimizers across all objectives, at each iteration. Iterative optimizers must balance \emph{exploiting} the knowledge gained from the evaluations with \emph{exploring} regions in the search space where the landscape is unknown and might hold better designs. This balance is particularly acute with limited experimental budgets, e.g., DNNs deployed on production or resource-constrained devices that can inhibit efficient finding of Pareto optimal designs. Recently, \camobo~\shortcite{abdolshah2019cost} proposed a cost-aware approach to identify the Pareto optimal designs by avoiding the designs with high evaluation costs in the design space. However, this can lead to aggressive exploitation behavior and generate sub-optimal designs. \pesmodec \ introduces a decoupled evaluation strategy where only a subset of objectives at any given location is evaluated. The decoupled evaluation provides significant improvements over a coupled evaluation, particularly when the experimentation budget is limited. However, selecting designs to evaluate without considering non-uniform evaluation costs can potentially lead to inefficient utilization of resources. 

To address these limitations, we propose the cost-aware decoupled MOBO approach \tool, which explicitly considers non-uniform objective evaluation costs and evaluates expensive objectives only if the gain of information is worth it. \tool \ extends the concepts of the state-of-the-art active learning algorithm \pal~\shortcite{zuluaga2013active} and \pesmodec~\shortcite{hernandez2015predictive}. To our knowledge, this is the first approach to propose a cost-aware decoupled evaluation strategy for MOBO. To formalize the notion of non-uniform evaluation costs of objectives, we define \emph{objective evaluation cost} in terms of computation time. Our acquisition function incorporates the uncertainty of the surrogate model’s predictions and objective evaluation costs to balance exploration
and exploitation and iteratively improve the quality of the Pareto optimal design space, also known as the \emph{Pareto
region}. It selects the objectives across which the design will be evaluated in addition to selecting the next design to evaluate. Consequently, we explicitly trade off the additional information
obtained through an evaluation with the cost of obtaining it, ensuring that we do not perform costly evaluations for little potential gain. By avoiding costly evaluations, we improve the efficiency of the search for Pareto optimal designs. We demonstrate \tool's promise through a comprehensive experimental evaluation on a range of different benchmarks. While we focus on DNNs, our proposed approach is general and can be extended to other applications. 



\subsection{Contributions} 
In summary, our contributions are as follows.

\begin{itemize}
    \item[$\blacktriangleright$] We propose \tool: a cost-aware approach for multi-objective Bayesian optimization that selects \emph{a design} and \emph{an objective} for evaluation. It allows to trade off the additional information gained through an evaluation and the cost being incurred as a result of the evaluation (Section \ref{sec:flexibo}).

    
     \item[$\blacktriangleright$] We comprehensively evaluate \tool \ on seven DNN architectures from three different domains and compare its performance to \parego~\shortcite{knowles2006parego}, \smsego~\shortcite{ponweiser2008multiobjective},  \pal~\shortcite{zuluaga2013active}, and \pesmo,\pesmodec~\shortcite{hernandez2016predictive}, and \camobo~\shortcite{abdolshah2019cost}. (Section~\ref{sec:evaluation}). The dataset and scripts to reproduce our findings are available at \url{https://github.com/softsys4ai/FlexiBO}.

\end{itemize}

\begin{figure}[h]
\centering
\subfloat[a][]{\includegraphics[width=0.48\textwidth]{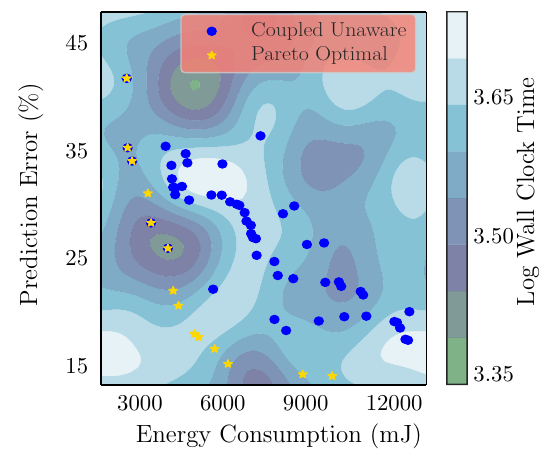} \label{fig:cuc_issue}} 
\subfloat[b][]{\includegraphics[width=0.48\textwidth]{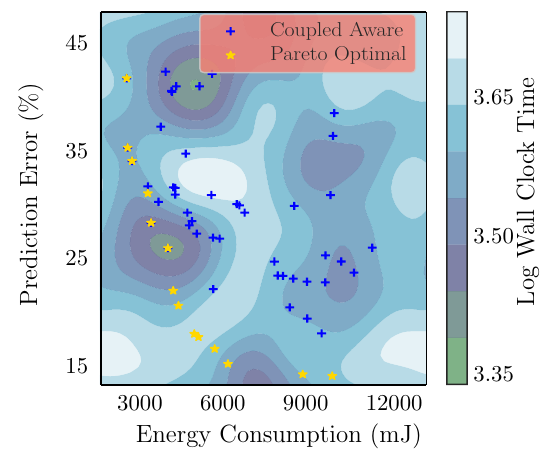} \label{fig:cac_issue}} \\
\subfloat[c][]{\includegraphics[width=0.48\textwidth]{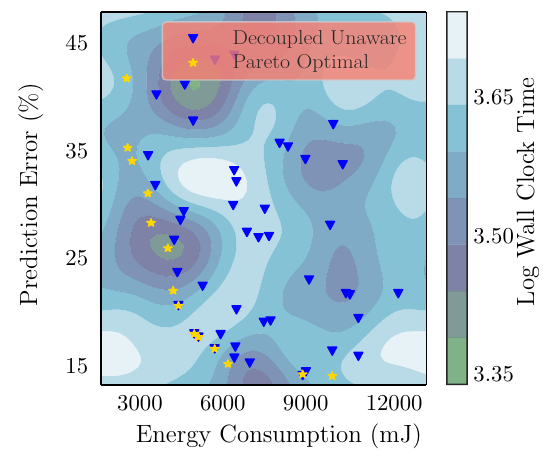} \label{fig:cud_issue}} 
\subfloat[d][]{\includegraphics[width=0.48\textwidth]{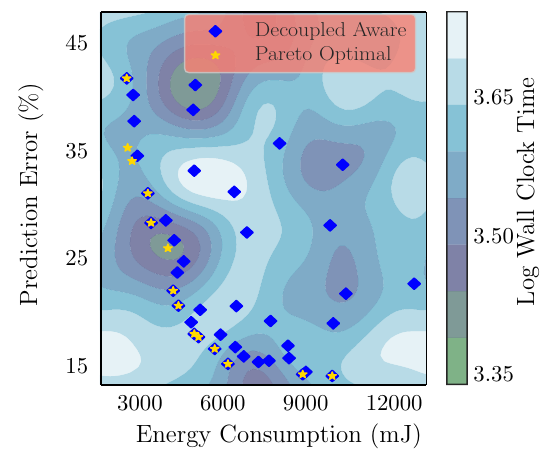} \label{fig:cad_issue}} 
 
  \caption{(a) Cost-unaware coupled approaches waste resources by evaluating designs with higher evaluation costs.  (b) Cost-aware coupled approaches can suffer poor performance if the Pareto optimal designs can only be found by evaluating objectives with high evaluation cost. (c) Cost-unaware decoupled approaches can invest a lot of resources in evaluating designs with low quality (high prediction error and high energy consumption) and do not perform well across both objectives. (d) Cost-aware decoupled approaches find designs with better quality compared to other approaches. Yellow stars indicates the location of Pareto optimal designs. The color indicates the evaluation cost (log wall-clock time) of a particular design.
}
\end{figure}
\section{Motivation}
\label{sec:motivate}
In this section, we discuss our motivation to propose a \emph{cost-aware} and \emph{decoupled} evaluation strategy. Based on the cost distribution assumptions and evaluation strategy, existing MOBO techniques can be subdivided into the following categories:  (I) cost-unaware coupled, (II) cost-aware coupled, and (III) cost-unaware decoupled approaches. Unlike cost-unaware approaches, cost-aware approaches assume the cost of evaluating different objectives is non-uniform. Decoupled approaches consider only a subset of objectives for evaluation at each iteration in the Bayesian optimization loop whereas coupled approaches evaluate all objectives. To show the clear advantage of our proposed cost-aware decoupled approach (to our knowledge, a gap in the MOBO literature that has not been addressed yet), we performed a sandbox experiment to optimize the prediction error and energy consumption of the image recognition DNN system SqueezeNet for the \textsc{CIFAR-10} dataset deployed on an \textsc{Nvidia JETSON TX2} device for inference on 5,000 test images. We use 8 \textsc{Nvidia Tesla K80} GPUs deployed on Google cloud for training with 45,000 training images. We tuned a small subset of design options from different layers of the system stack -- CPU frequency and GPU frequency from the hardware layer, swappiness from the operating system layer, memory growth from the  model compiler layer, and filter size, number of filters, and number of epochs from the DNN model layer. We use \pal \ as a cost-unaware coupled approach, \camobo \ as a cost-aware coupled approach,  \pesmodec \ as a cost-unaware decoupled approach, and \tool \  as a cost-aware decoupled approach. 

Cost-unaware coupled approaches are not sample (design) efficient for budget-constrained applications as they do not make the best utilization of resources by evaluating the selected designs across all objectives even for little or no gain. Figure~\ref{fig:cuc_issue} shows that most of the designs selected by the cost-unaware coupled approach are not close to the Pareto front and are concentrated in regions with high evaluation costs. It also has poor coverage of the objective space. 

\begin{figure}[h]
\centering
  \subfloat[a][]{\includegraphics[width=0.28\textwidth]{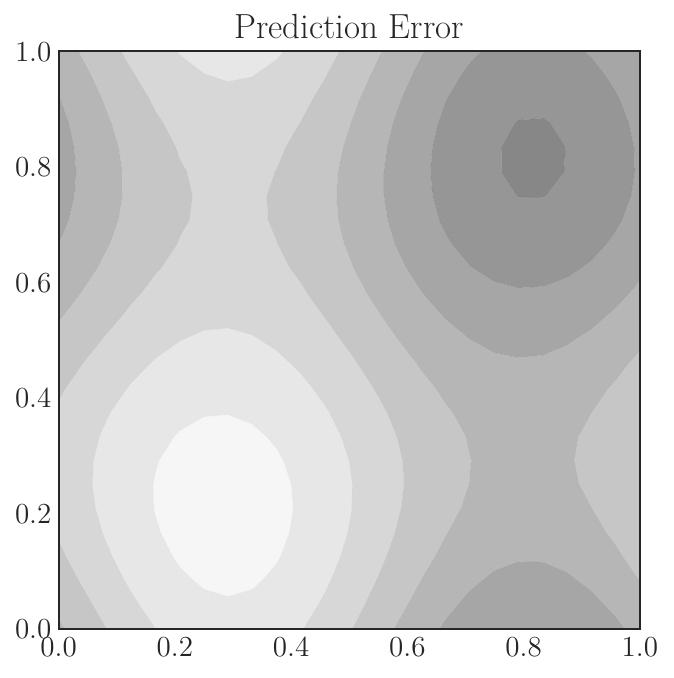} \label{fig:pesmo_err_issue}} 
  \subfloat[b][]{\includegraphics[width=0.28\textwidth]{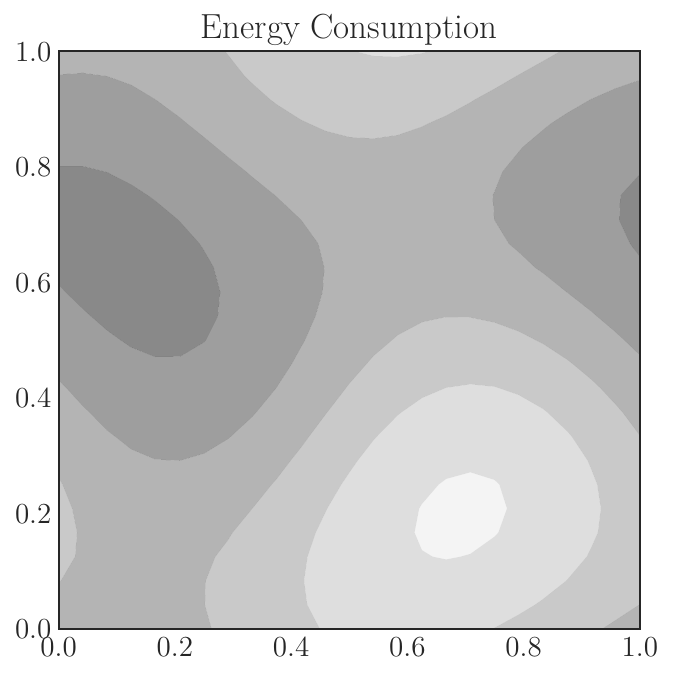} \label{fig:pesmo_energy_issue}} 
  \caption{Contour curves for (a) prediction error and (b) energy consumption of SqueezeNet by varying CPU Frequency and Number of Filters while keeping the other design options fixed. Decoupled unaware approaches perform poorly when objectives with different costs have the same complexity (both non-linear).
  } 
\end{figure}
Cost-aware coupled approaches consider different evaluation costs for different designs. As such approaches only evaluate cheap designs, good parts of the search space may be missed.  Figure~\ref{fig:cac_issue} shows that the cost-aware coupled strategy focuses on the cheap regions of the search space here and misses Pareto's optimal designs in expensive regions.

Cost-unaware decoupled approaches evaluate the more complex objectives a higher number of times than the less complex objectives. However, Figure~\ref{fig:pesmo_err_issue} and Figure~\ref{fig:pesmo_energy_issue} show that both the objectives (e.g., prediction error and energy consumption) have the same complexity. Cost unaware approaches are not particularly effective in such cases and produce suboptimal results as shown in Figure~\ref{fig:cud_issue}. This occurs as a result of them not making the best use of resources by evaluating a large number of low-quality designs (high prediction error and high energy consumption) for little information gain. 

 
Cost-aware decoupled approaches evaluate designs in any region if the information gain is large enough given the objective evaluation cost. In Figure~\ref{fig:cad_issue}, we observe that our cost-aware decoupled approach performs better than the other approaches and identifies more points on the Pareto front.
\section{Related Work}
We now discuss different directions of related work for multi-objective optimization.

\paragraph{Hardware-aware optimization of DNNs.}
One of the largest difficulties in producing energy-efficient DNNs is the
disconnect between the platform where the DNN is designed, developed, and
tested, and the platform where it will eventually be deployed and the energy it
consumes there~\shortcite{guo2017towards,chen2understanding,cai2017neuralpower,qi2016paleo,manotas2014seeds,sze2017efficient,chen2016eyeriss}.
Therefore, hardware-aware multi-objective optimization approaches have been introduced~\shortcite{zhu2018mobile,lokhmotov2018multi,cai2018proxylessnas,wu2019fbnet,whatmough2019fixynn} that enable automatic optimization of DNNs in the joint space of architectures, hyperparameters, and even the computer system stack~\shortcite{zela2018towards,iqbal_transfer_2019,nardi2019practical,hernandez2016designing}. Like these approaches,
\tool \ enables efficient multi-objective optimization in such joint configuration spaces.
Multi-objective neural architecture search (NAS) \shortcite{kim2017nemo,dong2018dpp,liu2017hierarchical} aims to optimize accuracy and limit resource consumption, for example by limiting the search space \cite{kim2017nemo}.
Several approaches characterize runtime, power, and the energy consumption of DNNs via analytical models, e.g., Paleo~\shortcite{qi2016paleo}, NeuralPower~\shortcite{cai2017neuralpower}, Eyeriss~\shortcite{chen2016eyeriss}, and Delight~\shortcite{rouhani2016delight}.  
However, they either rely on proxies like inference time for energy consumption or extrapolate energy values from energy-per-operation tables. They therefore cannot be used across different deployment platforms.

\noindent \paragraph{Multi-Objective Optimization with Different Acquisition Functions.}
There is a large body of research that identifies the complete Pareto front
using entropy-based acquisition functions. For example,
\mesmo \ \shortcite{wang2017max,belakaria2019max},
\mesmoc \ \shortcite{belakaria2020max}, and \pesmo \ \shortcite{hernandez2016predictive} determine the
Pareto front by reducing posterior entropy.
\smsego \ uses the maximum hypervolume improvement acquisition function to choose the next sample~\shortcite{ponweiser2008multiobjective}.
Different gradient-based multi-objective optimization algorithms
have been proposed to optimize
objectives more efficiently \shortcite{schaffler2002stochastic,desideri2012multiple}.
These methods were extended to use stochastic gradient descent
\shortcite{poirion2017descent,peitz2018gradient}. Active learning
approaches have been proposed to approximate the surface of the Pareto front
\shortcite{campigotto2013active} through the use of acquisition functions such
as expected hypervolume improvement~\shortcite{emmerich2008computation}
and sequential uncertainty reduction \shortcite{picheny2015multiobjective}.
Contemporary active learning approaches
like \pal \ and \epal \ tend to approximate the Pareto front
\shortcite{zuluaga2013active,zuluaga2016varepsilon} using the maximum diagonal of the uncertainties in the objective space as the acquisition function.
However, these methods do not take into account the varying
costs of the evaluations of the objective functions and are expensive.

\paragraph{Multi-Objective Optimization With Preferences.} Some methods
use \textit{preferences} in multi-objective optimization
with evolutionary methods
\shortcite{deb2006reference,thiele2009preference,kim2011preference}; although
these methods enable the user to guide the exploration of the design space of
systems~\shortcite{kolesnikov2019tradeoffs}, they are not sample-efficient,
which is essential for optimizing highly-configurable
systems~\shortcite{pereira2019learning,JC:MASCOTS16,JVKSK:SEAMS17,JVKS:FSE18,nair2018finding}, particularly for very large configuration spaces~\shortcite{acher2019learning}.
Recently, methods that use surrogate models for optimization with preferences
have been proposed~\shortcite{paria2018flexible,abdolshah2019multi}. These methods require the user to manually specify a preference though and are not cost-efficient.

\paragraph{Multi-Objective Optimization With Scalarizations.}
Different multi-objective optimization methods have been developed that use
scalarizations to combine multiple objectives into a single one such that
optimal solutions correspond to Pareto-optimal solutions. Examples include \textsc{ParEGO}, which uses random
scalarizations \shortcite{knowles2006parego}, weighted
product methods \shortcite{deb2001multi}, and utility functions
\shortcite{roijers2013survey,roijers2017interactive,roijers2018interactive,zintgraf2018ordered}. A major disadvantage of the scalarization approach is that without further assumptions (e.g., convexity) on the objectives, not all Pareto optimal solutions can be recovered. Therefore, solutions obtained by scalarization approaches tend to be sub-optimal.

\paragraph{Cost-Aware Multi-Objective Optimization Approaches.}
Recently, different cost-aware methods \shortcite{abdolshah2019cost,lee2020cost} have been proposed that incorporate the evaluation costs of objectives into account. They assign costs to designs in the design space and attempt to identity an optimal Pareto front by avoiding the costly designs; thereby selecting cheap designs for evaluation. These methods are either orthogonal or complimentary to our approach. \tool \ is a decoupled approach where we trade off the evaluation cost of an objective with the amount of information that can be gained.

\section{Background and Definitions}
\label{background}
In this section, we review MOBO and Pareto optimality and introduce the terminology and notation used in the rest of the paper. Table \ref{tab:notations} in the appendix lists the symbols and their descriptions used throughout the paper. 

\paragraph{Bayesian Optimization.} 
Bayesian Optimization (BO) is  an  efficient  framework  to  solve  global  optimization problems using black-box evaluations of expensive objective functions \cite{jones_efficient_1998}. Let $\mathfrak{X} \subset \mathbb{R}^d,$ where $d \in \mathbb{N}$, be a \emph{finite design space}. For single-objective Bayesian optimization (SOBO), we are given a real-valued objective function $f:\mathfrak{X} \rightarrow \mathbb{R}$, which can be evaluated at each design $\bm{x} \in \mathfrak{X}$ to produce an evaluation $y=f(\bm{x})$. Each evaluation of $\bm{x}$ is expensive in terms of the consumed resources. The main goal is to find a design $\bm{x}^* \in \mathfrak{X}$ that optimizes $f$ by performing a limited number of function evaluations. BO approaches use a cheap surrogate model learned from training data obtained using past function evaluations. They intelligently select designs for
evaluation by searching over the surrogate model, trading off exploration and exploitation to quickly direct the search towards an optimal design. 


\paragraph{Acquisition Function.} This is used to score the utility of evaluating a candidate design $\bm{x} \in \mathfrak{X}$ based on the statistical model. Some popular acquisition functions in the SOBO literature include expected improvement (EI)~\shortcite{emmerich2008computation}, upper confidence bound (UCB)~\shortcite{srinivas2012information}, predictive entropy search (PES)~\shortcite{hernandez2014predictive}, and max-value entropy search (MES)~\shortcite{wang2017max}. 

\begin{figure}[t]
  \begin{center}
    \includegraphics[width=\textwidth]{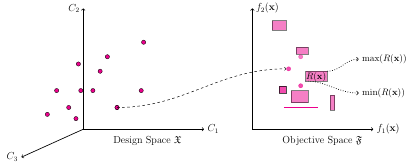}
    \end{center}
    \caption{The design space $\mathfrak{X}$ is mapped to the objective space $\mathfrak{F}$ for $n=2$. The objective space is showing examples uncertainty regions $R(\bm{x})$ for different designs $\bm{x}$.}
    \label{fig:mapping}
\end{figure}



\paragraph{MOBO.} 
In MOBO, the aim is to find a set of designs that simultaneously optimizes $n$ possibly conflicting objective functions $\bm{f}=f_1,\dots,f_n$, where $n \geq 2$ and ${f_i}:\mathfrak{X} \rightarrow \mathbb{R}$ for $1 \leq i \leq n$.
Each evaluation of a design $\bm{x} \in \mathfrak{X}$ produces a vector of
objective values $\bm{y}= (y_1, ..., y_n)$,
where $y_i=f_i (\bm{x})$ for $1 \leq i \leq n$. 

\paragraph{Pareto-Optimality.}
It is generally not possible to find a design that optimizes each objective equally, but instead, there is a trade-off between them. Pareto optimal designs represent the best compromises across all objectives. In the context of maximization, a design $\bm{x}$ is said to \emph{dominate}
another design $\bm{x}^\prime$, formally, $\bm{x} \succeq \bm{x}^\prime$ if $f_i(\bm{x}) \geq f_i(\bm{x}^\prime)$ for $1 \leq i \leq n$.
A design $\bm{x} \in \mathfrak{X}$ is called \emph{Pareto-optimal} if it is not
dominated by any other designs $\bm{x}^\prime \in \mathfrak{X}$, where $\bm{x}
\neq \bm{x}^\prime$. The set of designs $\mathfrak{X}^*$ is called the optimal
Pareto set and a \emph{hyperplane}\footnote{A subspace of the design space whose
dimension is one less than the design space.} passing through the corresponding
set of function values $\mathfrak{F}^*$ is called the \emph{Pareto front}. 

\paragraph{Surrogate Model} Surrogate models $\mathfrak{M_i}$ for $1 \leq i \leq
n$ are used to approximate the function to optimize, which is usually
computationally expensive to evaluate and not available in closed form.
Surrogate models are trained with evaluations of a small subset of the design
space $\mathfrak{X}$ and are used to predict the objective functions value using
$\hat{f}(\bm{x})=\bm{\mu}(\bm{x})$ with estimation uncertainty $\bm{\sigma}(\bm{x})$ for each design $\bm{x}$. 
The uncertainty region $R(\bm{x})$ of a design $\bm{x}$ is defined as a
hyper-rectangle of the width of the confidence region using $\bm{\mu}(\bm{x})$ and $\bm{\sigma}(\bm{x})$ (formally defined later).

\begin{figure}[htb]
  \begin{center}
    \includegraphics[width=.4\textwidth]{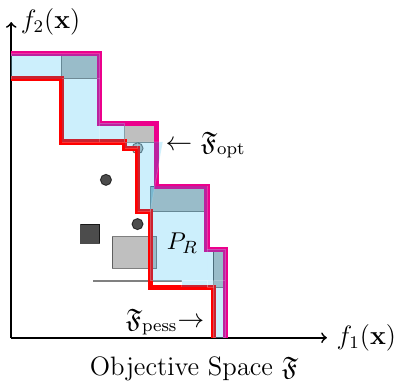}
    \end{center}
    \caption{The objective space is showing examples of Pareto fronts $\mathfrak{F}_{pess}$ and $\mathfrak{F}_{opt}$ passing through a subset of non-dominated designs. Here, non-dominated designs are shown in gray and dominated designs are shown in black. }
    \label{fig:pareto}
\end{figure}
\paragraph{Optimistic and Pessimistic Pareto front.} 
Each design $\bm{x}$ in the design space $\mathfrak{X}$ is assigned an uncertainty region $R(\bm{x})$ using the predictions of the objective functions $\bm{f}$ from the surrogate models $\mathfrak{M}$.
Figure \ref{fig:mapping} shows an example of uncertainty region of a design and its maximum value $\max(R(\bm{x}))$ and minimum value $\min(R(\bm{x}))$ for $n=2$ objectives. The maximum value of the uncertainty region $\max(R(\bm{x}))$ and minimum value of the uncertainty region $\min(R(\bm{x}))$ of a design $\bm{x}$ are regarded as the optimistic and pessimistic value of $\bm{x}$, respectively. A hyperplane passing through the non-dominated optimistic values of $\bm{x}$ is considered the optimistic Pareto front $\mathfrak{F}_{opt}$. Similarly, a pessimistic Pareto front $\mathfrak{F}_{pess}$ is constructed by a hyperplane passing through the pessimistic values of $\bm{x}$.  

\paragraph{Pareto Region.} The region bounded by the optimistic Pareto front
$\mathfrak{F}_{opt}$ and pessimistic Pareto front $\mathfrak{F_{pess}}$ is
defined as the Pareto region $P_R$ (shown as the blue shaded region in Figure~\ref{fig:pareto}).


\paragraph{Objective evaluation cost.} Objective evaluation cost $\theta_i(\bm{x})$ of a design $\bm{x}$ is the computational effort required to evaluate design $\bm{x}$ for an objective $f_i$. 


\section{\tool: \underline{Flexi}ble  Multi-Objective \underline{B}ayesian \underline{O}ptimization}
\label{sec:flexibo}
In this section, we explain the technical details of our proposed  \underline{Flexi}ble  Multi-Objective \underline{B}ayesian \underline{O}ptimization (\tool) algorithm. \tool \ aims to identify the optimal Pareto front $\mathfrak{F}^*$
by evaluating a small subset of designs in the design space $\mathfrak{X}$ that uses a
cost-aware acquisition function to incorporate the evaluation costs of each
objective in the standard Bayesian optimization framework. Given the same
budget $\theta_T$, the cost-awareness of the acquisition function enables \tool \ to sample
the search space more efficiently compared to other state-of-the-art approaches. 

\subsection{Algorithm Design}

\tool \ is an active learning algorithm that selects a sequence of designs $(\bm{x}_1,...,\bm{x}_T)$ in the
design space $\mathfrak{X}$ for evaluation to determine the Pareto-optimal designs; the designs classified as Pareto-optimal are then
returned as the prediction $\mathfrak{\hat{F}}^*$ for $\mathfrak{F}^*$. Rather than evaluating each design $\bm{x}$ against all objectives $f_i$ for $1 \leq i \leq n$, our cost-aware approach
iteratively evaluates a selected design $\bm{x}$ only across the most informative
objective. \tool \ evaluates a design $\bm{x}$ \ across an objective $f_i$ if
the change in hypervolume of the Pareto region is large enough compared to the
objective evaluation cost $\theta_i$. This allows \tool \ to avoid expensive
measurements for little or no change in hypervolume and to only evaluate across
an objective when the change in hypervolume is worthy compared to the evaluation cost. Formally, \tool \ is a cost-aware multi-objective optimization approach that iteratively and adaptively selects a sequence of designs and objectives $((\bm{x}_1,f_{1,i}),...,(\bm{x}_T,f_{T,i}))$ for $1 \leq i \leq n$ across which the selected designs are evaluated to predict the Pareto front $\mathfrak{\hat{F}}^*$.


We then fit a separate surrogate model $\mathfrak{M_i}$ for each objective function $f_i$ for $1 \leq i \leq n$. We select m designs set $\mathfrak{X_m}$ from the design space $\mathfrak{X}$ using Monte-Carlo sampling~\cite{shapiro2003monte}.
 The objective values of a design $\bm{x}$ that has not been evaluated across any objective
are estimated by
$\bm{\hat{f}}(\bm{x})=\bm{\mu}(\bm{x})=(\mu_1(\bm{x}),\dots,\mu_n(\bm{x}))$, and
the associated uncertainty is estimated by
$\bm{\sigma}(\bm{x})=(\sigma_1(\bm{x}),\dots,\sigma_n(\bm{x}))$. If a design $\bm{x}$ is evaluated across an objective $f_i$, the associated uncertainty is zero against $f_i$.
At each iteration $t$, we use the $\bm{\mu}_t(\bm{x})$ and $\bm{\sigma}_t(\bm{x})$ values to determine the uncertainty region $R_t(\bm{x})$ for each design $\bm{x} \in \mathfrak{X_m}$. We define the uncertainty region associated with a prediction of the surrogate model as follows:
\begin{equation}
\label{eq:uncertainty}
R_{t}(\bm{x})=\{ \bm{y}: \bm{\mu}_t(\bm{x})-\sqrt{\beta_{t}} \bm{\sigma}_t (\bm{x}) \leq \bm{y} \leq  \bm{\mu}_t(\bm{x})+\sqrt{\beta_{t}} \bm{\sigma}_t (\bm{x}) \},
\end{equation}
where $\beta_t$ is a scaling parameter that controls the
exploration-exploitation trade-off.
Similar to \pal \ \shortcite{zuluaga2013active,zuluaga2016varepsilon}, we use $
\beta_t= 2/9 \log(n|\mathfrak{X_m}|\pi^2t^2/{6\delta})$ for $\delta \in (0,1)$.The
dimension of $R_{t}(\bm{x})$ depends on the number of objectives $n$. Later, we
exploit the information about the uncertainty regions to determine the
non-dominated designs set $\mathfrak{U}$.
We then use the optimistic and
pessimistic values of the non-dominated designs in $\mathfrak{U}$ to build the
optimistic Pareto front $\mathfrak{F}_{opt}$ and pessimistic Pareto front
We now employ our cost-aware acquisition function, which makes use of an information gain based on objective space entropy. Being cost-aware, our
proposed acquisition function $\alpha_{t,i}(\bm{x})$ considers the evaluation cost $\theta_{t,i}$ across each objective $f_i$:
\begin{align}
      \alpha_{t,i}(\bm{x}) & = \frac {\Delta V(\{\bm{x},f_{t,i}(\bm{x})\}, \mathfrak{\hat{F}}^* | \mathfrak{X_m}^*)}{\theta_{t,i}} \label{eq:6}\\
      & = \frac{V(P_R | \mathfrak{\hat{F}}^*) -
      V\left(P_R|\mathfrak{\hat{F}}^*_{R_{t,i}(\bm{x})=\bm{\mu}_{t,i}(\bm{x})}\right)}{\theta_{t,i}} \label{eq:7}\\
      & = \frac{\Delta{V_{t,i}}}{{\theta_{t,i}}}
\end{align}

Here, $\alpha_{t,i}(\bm{x})$ computes the amount of information that can be gained per cost for a design $\bm{x}$ to be evaluated for an objective $f_i$. In Equation \ref{eq:7}, we compute the gain of information as the change of volume of the Pareto region if the Pareto front $\mathfrak{\hat{F}}^*= \mathfrak{F}_{opt} \cup \mathfrak{F}_{pess}$ is updated by setting the
uncertainty values $R_{t,i}(\bm{x})$ of $\bm{x}$ to its mean $\mu_{t,i}(\bm{x})$ for the corresponding designs in $\mathcal{X}_{m}^*$.
Our acquisition function computes the change of volume $\Delta V_{t,i}$ of the Pareto region $P_R$ across each objective $f_i$ to judiciously determine the gain of information that would be achieved if design $\bm{x}$ is evaluated for $f_i$. We select a design $\bm{x}_t$ and an objective $f_{t,i}$ using the following:
\begin{align}
\bm{x}_{t}, f_{t,i}=\text{argmax}_{\bm{x} \in  \mathcal{X}_{m}^* \ \text{for each} \ f_i} \ \alpha_{t,i}(\bm{x})
\end{align}
Here, we identify the most promising design for an objective function that gains the most information given the cost of evaluating it. Finally, we update the
surrogate model $\mathcal{M}_i$ corresponding to the chosen objective function $f_i$ by incorporating the newly-evaluated design and objective value. We stop when the maximum budget $\theta_{T}$ is exhausted or the maximum change of volume of the Pareto region becomes zero (indicating that all designs in the Pareto region have been evaluated for each objective ), whichever occurs earlier. Finally, we return the Pareto front obtained.  
Every iteration $t$ consists of three stages: (1) modeling, (2) construction of
the Pareto region, and (3) sampling. To initialize \tool, we evaluate $N_0$
samples for each objective $f_i$ to and populate corresponding evaluated designs
set $S_i$ for $1 \leq i \leq n$. We also determine the average computational
effort $\theta_i$ for each objective $f_i$ before proceeding to the iterative
procedure. We outline the pseudocode for the \tool \ implementation in Algorithm~\ref{alg:flexibo}.  

\begin{figure}
 
  \begin{center}
    \includegraphics[width=0.4\textwidth]{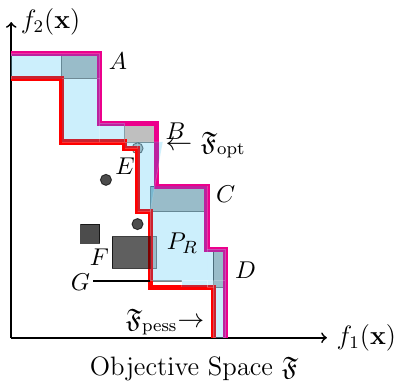}
  \end{center}
     
    \caption{Example showing pruning of non-dominated points to construct $\mathfrak{F}_{opt}$ and $\mathfrak{F}_{pess}$. }
    
    \label{fig:pareto-cases}
\end{figure}



\begin{algorithm}[t]
\setcounter{AlgoLine}{0}
\KwInput{design space $\mathfrak{X}$; maximum budget $\theta_{T}$; number of initial designs $N_0$; }
%
\rule{0.35\textwidth}{1pt} \emph{Initialization} \rule{0.34\textwidth}{1pt}

$S_i$ = Evaluate $N_0$ designs for each objective $f_i$

Determine average computational effort $\theta_{i}=(\sum_{t=1}^{N_0}\theta_{t,i})/N_0$ for each $f_i$

$t$ = $N_0$ and $\theta = 0$



\While{$\theta \leq \theta_{T}$}{
\rule{0.3\textwidth}{1pt} \emph{Modeling} \rule{0.3\textwidth}{1pt}

Train surrogate models $\mathfrak{M}_{t,i}$ using corresponding evaluated designs set $S_{i}$ for $f_i$


Obtain $\bm{\mu}_{t}(\bm{x})=(\mu_{t,i}(\bm{x}))_{1 \leq i \leq n}$ and $\bm{\sigma}_{t}(\bm{x})=(\sigma_{t,i}(\bm{x}))_{1 \leq i \leq n}$ using $(\mathfrak{M}_{t,i})_{1 \leq i \leq n}$ for all $\bm{x} \in \mathfrak{X_m}$. 

Compute uncertainty region $R_t(\bm{x})$ of each design $\bm{x} \in \mathfrak{X_m}$ using Equation~\ref{eq:uncertainty} 

\rule{0.21\textwidth}{1pt} \emph{Pareto region construction} \rule{0.20\textwidth}{1pt}

 $\mathfrak{U}=\varnothing$ 
 
\For {all  $\bm{x} \in \mathfrak{X_m}$}{
    \If {no $\bm{x} \neq \bm{x}^\prime$ for $\bm{x}^\prime \in \mathfrak{X_m}$ exists  such that $\min(R_t(\bm{x})) \succeq \max(R_t(\bm{x}^\prime))$}{
         $\mathfrak{U}=\mathfrak{U} \cup \{ \bm{x} \}$} 
     }

Identify optimistic Pareto front $\mathfrak{F}_{opt}$ and pessimistic Pareto front $\mathfrak{F}_{pess}$  using $\mathfrak{U}$ and $R_t(\bm{x})$ with Equation~\ref{opt}, Equation~\ref{pess-1}, and Equation~\ref{pess-2} 


\rule{0.3\textwidth}{1pt} \emph{Sampling} \rule{0.3\textwidth}{1pt}

Compute acquisition function $\alpha_{t,i}(\bm{x})$ across each objective $f_i$ using $\bm{x} \in \mathfrak{X_m}^*$ with Equation~\ref{eq:7} 

Update $t$ = $t+1$

Choose the next sample $\bm{x}_{t}$ and objective $f_{t,i}$ using Equation~\ref{eq:max-acq}

Evaluate $\bm{x}_t: y_{t,i} = f_{t,i}(\bm{x}_t)$ 

Aggregate data $S_{i}=S_i \cup \{(\bm{x}_t,y_{t,i})\}$

Update $\theta_{t,i}= ((t-1)*\theta_i + \theta_{t,i})/{t}$

Update $\theta =\theta + \theta_{t,i}$ 

}
\textbf{return} Return the non-dominated designs from the evaluated designs set $(S_i)_{1 \leq i \leq n}$ as the Pareto front using Equation~\ref{eq:pf}

\caption{The \tool \ algorithm.
}
\label{alg:flexibo}
\end{algorithm}


\noindent \subsubsection{Modeling}
At each iteration $t$, we train a surrogate model $\mathfrak{M}_i$ using the
samples in the evaluated designs set $S_i$ for objective $f_i$. As \tool \ selects one
objective $f_i$ for evaluation per iteration, only the surrogate model $\mathfrak{M}_i$ corresponding to the selected objective $f_i$ needs to be updated. Then, we determine $\mathfrak{X}_m$ from $\mathfrak{X}$ using Monte-Carlo sampling~\cite{shapiro2003monte}. At this point, we determine the uncertainty region $R_t(\bm{x})$ of each design $\bm{x} \in \mathfrak{X_m}$ using Equation~\ref{eq:uncertainty}. The 2-dimensional objective space $\mathfrak{F}$ in Figure \ref{fig:mapping} is showing examples of uncertainty regions for different designs $\bm{x}$.
As shown in Figure \ref{fig:mapping}, the uncertainty region
$R_t(\bm{x})$ of a design $\bm{x}$ that is not evaluated across any of the two
objectives, $f_1$ and $f_2$, is a rectangle. If $\bm{x}$ is evaluated across one objective, say $f_2$, the uncertainty across $f_2$ will be eliminated (assuming measurements
contain no noise) and $R_t(\bm{x})$ will become a line across $f_1$. Once, $\bm{x}$
is evaluated across both objectives, $R_t(\bm{x})$ is expressed by a
point (indicating no uncertainty across $f_1$ and $f_2$). 


\noindent \subsubsection{Pareto Region Construction}

After the uncertainty region $R_t(\bm{x})$ for each design $\bm{x} \in \mathfrak{X_m}$, we identify the set of non-dominated designs $\mathfrak{U}$ using the following rule:
\begin{equation}
   \label{eq:pf}
   \begin{split}
    \bm{x} \in \mathfrak{U}  \ \text{if} \ \min(R_t(\bm{x})) \succeq \max(R_t(\bm{x}^\prime))
    \ \text{for} \ \bm{x} \neq \bm{x}^\prime \ \text{and} \ \bm{x}, \bm{x}^\prime \in \mathfrak{X_m}  
    \end{split}
\end{equation}
Figure \ref{fig:pareto} shows examples of non-dominated designs (gray color) and dominated designs (black color) in the objective space for $n=2$.
Next, we identify the set of Pareto-optimal solutions $\mathfrak{X_m}^*$ and Pareto front $\mathfrak{\hat{F}}^*=\{ {\mathfrak{F}_{opt}} \cup \mathfrak{F}_{pess}   \}$
for the purpose of constructing the Pareto region $P_R$ by pruning designs in $\mathfrak{U}$. A design $\bm{x} \in \mathfrak{U}$ is only included in $\mathfrak{F}_{opt}$ if the optimistic value $\max(R_t(\bm{x}))$ of $\bm{x}$ is not dominated by the optimistic value $\max(R_t(\bm{x}^\prime))$ of another design $\bm{x}^\prime$ across all objectives as follows.
\begin{equation}
   \label{opt}
   \begin{split}
    \bm{x} \in \mathfrak{F_{opt}}  \ \text{if} \ \max(R_t(\bm{x})) \succeq
    \max(R_t(\bm{x}^\prime)) \ \text{for} \ \bm{x} \neq \bm{x} \ \text{and} \ \bm{x}, \bm{x}^\prime \in \mathfrak{U}  
    \end{split}
\end{equation}
Figure \ref{fig:pareto-cases} shows an example where non-dominated designs $F$ or $G$ are not included in $\mathfrak{F}_{opt}$ as the optimistic values of $F$ or $G$ are dominated by the optimistic values of non-dominated designs $B$ or $C$. 

We directly add the pessimistic value $\min(R_t(\bm{x}))$ of a design $\bm{x}$ to $\mathfrak{F}_{pess}$ if it remains non-dominated by the pessimistic value $\min(R_t(\bm{x}^\prime))$ of any other point $\bm{x}^\prime$ as follows.
\begin{equation}
   \label{pess-1}
   \begin{split}
    \bm{x} \in \mathfrak{F_{pess}}  \ \text{if} \ \min(R_{t}(\bm{x})) \succeq
    \min(R_t(\bm{x}^\prime)) \ \text{for} \ \bm{x} \neq \bm{x}^\prime \in \mathfrak{U} 
    \end{split} 
\end{equation}
As shown in Figure \ref{fig:pareto-cases}, pessimistic values of $B$, $D$, $E$ etc.\ are added to $\mathfrak{F}_{pess}$ using the above rule in Equation \ref{pess-1}. However, the
uncertainty regions $R_t(\bm{x}^\prime)$ of some designs $\bm{x}^\prime \in \mathfrak{U}$ ruled out of $\mathfrak{F}_{pess}$ using Equation \ref{pess-1} can have some degree of overlap with the uncertainty region $R_t(\bm{x})$ of a design $\bm{x} \in \mathfrak{F}_{pess}$. Consider the uncertainty regions of $F$ and $G$ in Figure
\ref{fig:pareto-cases}. Though their pessimistic values are dominated by the pessimistic values of $B$ and $C$, there is some overlap of the uncertainty regions of $F$ and $G$ with the uncertainty regions of $B$ and $C$ across an objective, in this case $f_1$. Overlapping uncertainty regions of $F$ and $G$ with $C$ are shown in gray as they remain non-dominated by the pessimistic value of $C$ in Figure~\ref{fig:pareto-cases}. In such cases, the pessimistic values of $F$ and $G$ are updated with the minimum values of the overlapping non-dominated uncertainty region using the following rule: 
\begin{equation}
   \label{pess-2}
   \begin{split}
    \min(R_{t,i}(\bm{x}^\prime))=\min(R_{t,i}(\bm{x}))\ \text{if} \ \min(R_t(\bm{x})) \succeq
    \min(R_t(\bm{x}^\prime)) \ \text{and}\ \\
    \min(R_{t,i}(\bm{x})) \nsucceq \max(R_{t,i}(\bm{x}^\prime)) \ \text{for each} \ f_i \ \text{where,} \ \bm{x} \neq \bm{x}^\prime \ \text{and} \ \bm{x},\bm{x}^\prime \ \in \mathfrak{U} 
    \end{split} 
\end{equation}
Later, updated pessimistic values of $F$ or $G$ are added to the pessimistic
Pareto front  $\mathfrak{F}_{pess}$ if it remains non-dominated. Note that there can be more than one design in $\mathfrak{F}_{pess}$ whose pessimistic value can dominate the pessimistic value of another design not yet included in $\mathfrak{F}_{pess}$ and has an overlap. We need to repeat the above process for each of those designs and finally update to a value that remains non-dominated. This process of identifying $\mathfrak{F}_{pess}$ ensures that any design that has the potential to be included in the pessimistic Pareto front is not discarded from our consideration.
Finally, the Pareto region $P_R$ bounded by $\mathfrak{F}_{opt}$ and $\mathfrak{F}_{pess}$ is constructed.  

\begin{figure}[t]
    \centering
    \includegraphics[width=\textwidth]{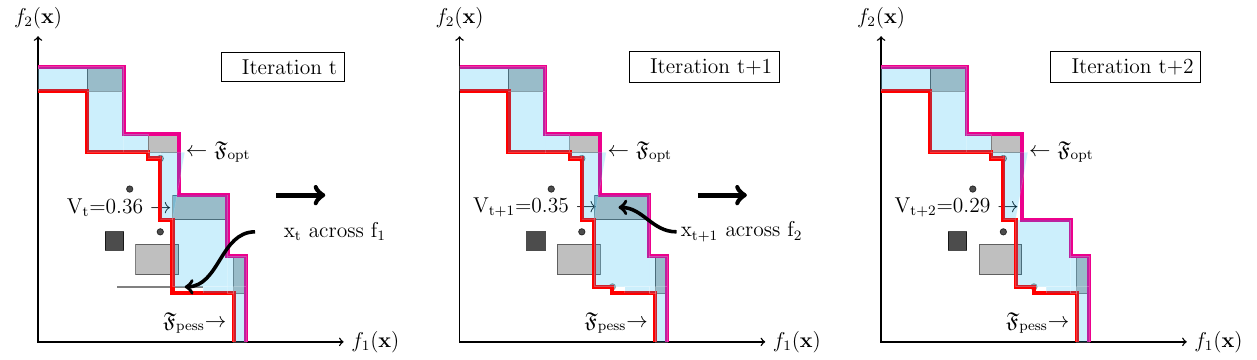}
    \caption{Decrease of the volume of the Pareto region with each iteration.
    }
    \label{fig:vol}
\end{figure}

\noindent \subsubsection{Sampling}
\label{sec:acquisition}

At this stage, we select the next design $\bm{x}_{t}$ and objective $f_{t,i}$ for evaluation using our proposed acquisition function $\alpha_{t,i}(\bm{x})$ by the following:
\begin{equation}
    \label{eq:max-acq}
    \bm{x}_{t}, f_{t,i}=\text{argmax}_{\bm{x} \in \mathfrak{X_m}^* \ \text{for each} \ f_i} \ \alpha_{t,i}(\bm{x}) 
\end{equation}

Here, we only use the designs in the Pareto optimal set $\mathfrak{X_m}^*$ whose function values constitute the Pareto fronts $\mathfrak{F}_{opt}$ and $\mathfrak{F}_{pess}$ in our acquisition function calculation.
We exclude designs not located on the hyperplanes passing through $\mathfrak{F}_{opt}$ and $\mathfrak{F}_{pess}$ as they do not contribute to the change of the volume of Pareto region $P_R$ when their uncertainty across any objective is reduced to zero.
Intuitively, this helps us to speed up computation. 

At every iteration, the evaluated design leads to a decrease in the volume of the Pareto region, as illustrated in Figure~\ref{fig:vol}.
We update the set of evaluated
designs $S_i$
for $f_{t,i}$ by adding $\{\bm{x}_{t},f_{t,i}(\bm{x}_{t})\}$.
Additionally, we update the computational effort $\theta_i$ for objective
$f_{t,i}$ using $\theta_i= ((t-1)*\theta_i + \theta_{t,i})/{t}$, where
$\theta_{t,i}$ is the computational effort to evaluate $\bm{x}_t$ across
$\bm{f}_{t,i}$.
Once the maximum budget $\theta_{T}$ is exhausted, \tool \ returns the non-dominated designs as approximate Pareto front using the evaluated designs $\bm{x} \in S_i$ for each objective $f_i$. Note that the estimated mean $\mu_i(\bm{x})$ is used as the objective value for $f_i$ if a design $\bm{x} \in S_i$ is not evaluated across $f_i$ while determining the Pareto front. 


\section{Evaluation}
\label{sec:evaluation}
\begin{figure}
    \centering
    \includegraphics[width=0.9\textwidth]{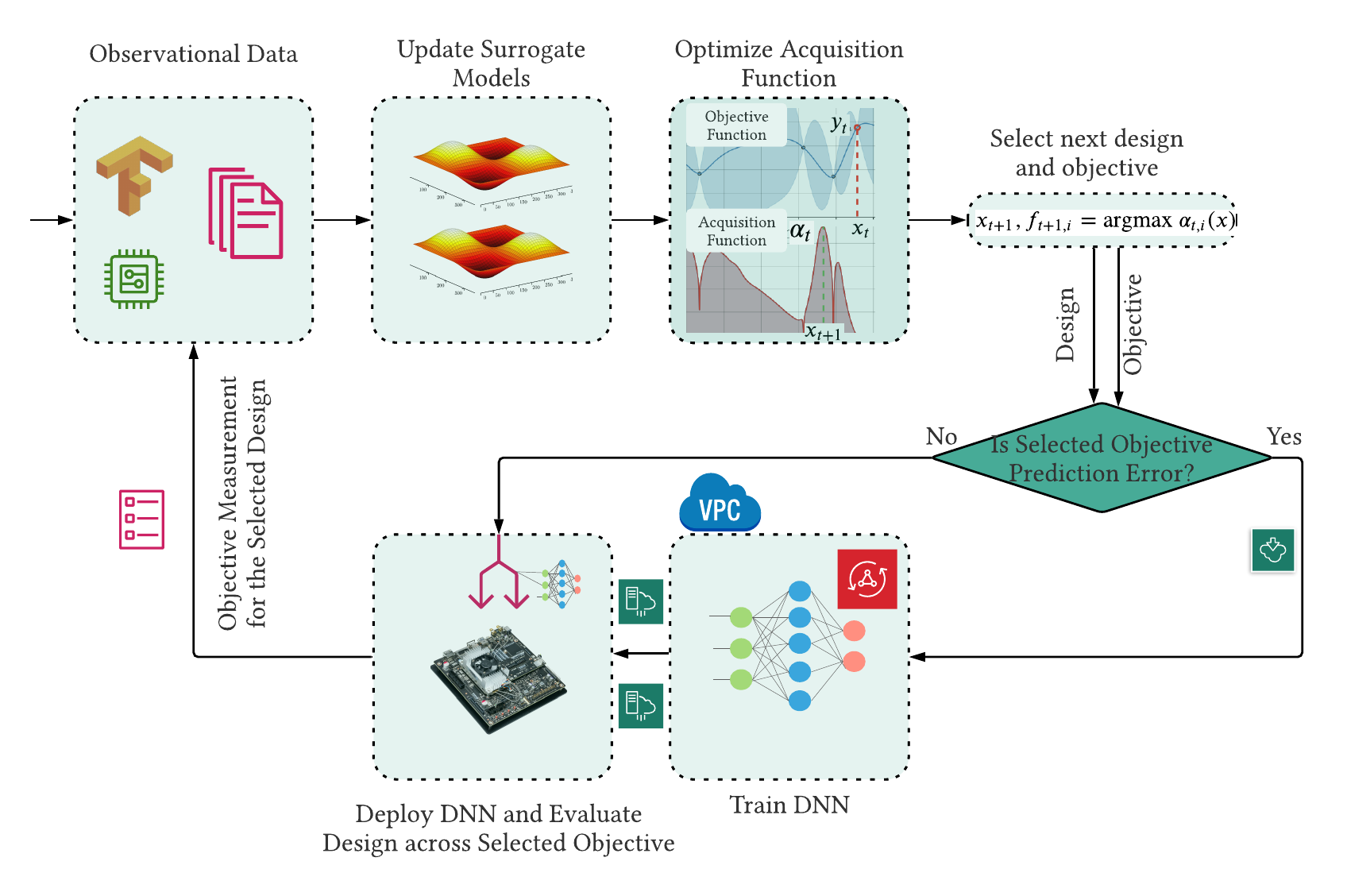}
    \caption{Experimental Setup used for \textsc{\tool }.
    }
    \label{fig:implementation}
\end{figure}
In this section, we evaluate the following research questions (RQs): 
\begin{itemize}
    \item[$\blacktriangleright$] \textbf{RQ1}: How to select the objective evaluation function for \tool \ to optimize multiple objectives for DNNs?
    \item[$\blacktriangleright$] \textbf{RQ2}: How effective is \tool \ in comparison to state-of-the-art multi-objective optimization approaches for
     \begin{itemize}
        \item[$\checkmark$] different DNNs of different applications?
         \item[$\checkmark$] different DNNs of varying sizes (e.g., number of hyperparameters)?
     \end{itemize}
     \item[$\blacktriangleright$] \textbf{RQ3}: How sensitive is \tool \ when different surrogate models are used?

\end{itemize}

\begin{table*}[t]
\centering
\caption{The DNN architectures and datasets used in the experimental evaluation.}
\resizebox{\linewidth}{!}{\begin{tabular}{llllllllll}

\doubleRule
\textsc{Domain}&\textsc{Architecture}&\textsc{Dataset}&\textsc{Compiler} & \textsc{Num. Layers} & \textsc{Num. Params} &\textsc{Train Size}& \textsc{Test Size}\\

\hlineB{2}

\multirow{5}{*}{\textsc{Image}}&\textsc{Xception}&\textsc{ImageNet}& Keras & 71&22M&100K & 10K \\
&\textsc{MobileNet}&\textsc{ImageNet}& Keras& 28&4.2M&100K & 10K\\
&\textsc{LeNet}&\textsc{MNIST}& Keras& 7&60K&50K & 10K\\
&\textsc{ResNet}&\textsc{CIFAR-10}& Keras& 50&25M&45K & 5K\\
&\textsc{SqueezeNet}&\textsc{CIFAR-10}&Keras& 3&1.2M&45K & 5K\\
\hlineB{2}

\multirow{2}{*}{NLP}&\textsc{Bert}&\textsc{SQuAD 2.0}&PyTorch& 12&110M&56K & 5K\\
&\textsc{Bert}&\textsc{IMDB Sentiment}&PyTorch& 12&110M&25K & 2K\\
\hlineB{2}

\multirow{1}{*}{\textsc{Speech}}&\textsc{Deepspeech}&\textsc{Common Voice}&PyTorch& 9&68M&300 (hrs) & 2 (hrs)\\
\doubleRule
\end{tabular}}
\label{tab:datasets}
\end{table*}

\subsection{Experimental Setup}
We discuss the baselines, datasets, and experimental setup to evaluate \tool in this section.

\subsubsection{Baselines}
We compare \tool \ to the following baselines:

\noindent \paragraph{\pesmo, \pesmodec~\shortcite{hernandez2015predictive}.} These methods employ an acquisition function based on input space entropy and iteratively select the design that maximizes the information gained about the optimal Pareto set. Both of these methods are cost-aware, with pesmo employing a coupled evaluation strategy and pesmodec employing a decoupled evaluation strategy. 

\noindent \paragraph{\pal~\shortcite{zuluaga2013active}.} An active learning algorithm that samples the design space by classifying designs as Pareto optimal or not to identify the Pareto front. This method uses a cost unaware coupled evaluation strategy.

\noindent \paragraph{\parego~\shortcite{knowles2006parego}.} Transforms the multi-objective problem into a single-objective problem using a scalarization technique.

\noindent \paragraph{\smsego~\shortcite{ponweiser2008multiobjective}.}  This method is given by the gain in hyper-volume obtained by the corresponding optimistic estimate, after an $\epsilon$-correction has been made. The hypervolume is simply the volume of points in functional space above the Pareto front (this is simply the function space values associated to the Pareto set), with respect to a given reference.
     
\noindent \paragraph{\camobo~\shortcite{abdolshah2019cost}.} The acquisition function in \camobo \ uses Chebyshev scalarization for objective functions to ensure the solutions satisfy Pareto optimality, and a cost function as a component of the acquisition function that incorporates the user’s prior knowledge of the search space. This multi-objective optimization method uses a cost aware coupled evaluation scheme.


We run each optimization pipeline $5$ times using different initial evaluations, where the initial evaluations in one run are the same for all methods.

\subsubsection{Datasets}
We use seven DNN architectures from three different problem domains;
\textsc{Image}, \textsc{NLP}, and \textsc{Speech}. For each architecture, we select the most common dataset and compiler typically used in practice. Table \ref{tab:datasets} lists the architectures, datasets, compilers, and the sizes of the training and test sets used in our experiments. 

\noindent \paragraph{\textsc{Image}.}
To evaluate the performance of \tool \ for image recognition applications, we use the Xception~\shortcite{chollet2017xception}, MobileNet~\shortcite{sandler2018mobilenetv2}, LeNet~\shortcite{lecun2015lenet}, ResNet~\shortcite{he2016deep}, and SqueezeNet~\shortcite{iandola2016squeezenet} architectures. For both Xception and MobileNet, we use the ImageNet ILSVRC2017 challenge dataset~\shortcite{russakovsky2015imagenet} and randomly select 100,000 train and 10,000 test images for our experiments. We use the MNIST dataset~\shortcite{lecun-mnisthandwrittendigit-2010} of handwritten images for LeNet. Our training and test datasets consist of 45,000 and 5,000 images, respectively. For our evaluation of \tool \ on ResNet and SqueezeNet, we use the CIFAR-10 dataset~\shortcite{krizhevsky2009learning}, which consists of 60,000 images of size 32$\times$32 with 10 classes (6,000 images per class). We use 50,000 images for training and the remaining 10,000 images for testing. 

\noindent \paragraph{NLP.}
We use the popular BERT~\shortcite{devlin2018bert} architecture for our
evaluation of \tool \ for NLP applications. We combine BERT on 2 benchmark
datasets: a question answering dataset, SQuAD
2.0~\shortcite{rajpurkar2016squad}, and the IMDB Movie Review Sentiment Analysis datatset. Out of 130,319 training  and 8,863 testing examples of the original SQuAD 2.0 dataset, we randomly select
56,000 training and 5,000 testing examples for our experiments with BERT (termed
BERT-SQuAD). For the IMDB movie review dataset
(termed BERT-IMDB), we use all 25,000 binary sentiment analysis training examples for training and randomly select
2,000 examples for testing out of the 25,000 testing examples provided in the IMDB dataset. 

\noindent \paragraph{\textsc{Speech.}}
To evaluate the performance of \tool \ for speech recognition, we use
DeepSpeech~\shortcite{hannun2014deep} with the Common Voice dataset~\shortcite{commonvoice}.
We randomly extract 300 hours of voice data for 5 different languages (English,
Arabic, Chinese, German, and Spanish) from nearly 3,700 hours of voice data of the Common Voice dataset for training. To evaluate the prediction error we test on 2 hours of voice data.


\begin{table}[h]

\caption{DNN-specific design options and their values.}

\resizebox{\textwidth}{!}{\begin{tabular}[t]{lll}
\doubleRule
\textsc{Architecture}&\textsc{Design Option}& \textsc{Value/Range} \\
\hlineB{2} 
\multirow{5}{*}{\textsc{Xception}}&Number of Filters Entry Flow &16, 32, 64, 128, 256 \\
&Number of Filters Middle Flow &16, 32, 64, 128, 256\\
&Filter Size Entry Flow &(1$\times$1), (3$\times$3), (5$\times$5), (7$\times$7), (9$\times$9)\\
&Filter Size Middle Flow &(1$\times$1), (3$\times$3), (5$\times$5), (7$\times$7), (9$\times$9)\\
&Filter Size Exit Flow &(1$\times$1), (3$\times$3)\\

\hlineB{2}
\multirow{6}{*}{\textsc{MobileNet}}&Number of Filters Stem &16, 32, 64, 128, 256 \\
&Filter Size Stem &(1$\times$1), (3$\times$3), (5$\times$5), (7$\times$7), (9$\times$9)\\
&Number of Filters Depthwise Block One &16, 32, 64, 128, 256, 512, 1024\\
&Number of Filters Depthwise Block Two &16, 32, 64, 128, 256, 512, 1024\\
&Number of Filters Depthwise Block Three &16, 32, 64, 128, 256, 512, 1024\\
&Number of Filters Depthwise Block Four &16, 32, 64, 128, 256, 512, 1024\\

\hlineB{2}
\multirow{8}{*}{\textsc{LeNet}}&Number of Filters Layer 1&16, 32, 64, 128, 256, 512, 1024\\
&Filter Size Layer 1&(1$\times$1), (3$\times$3), (5$\times$5), (7$\times$7), (9$\times$9)\\
&Number of Filters Layer 2&16, 32, 64, 128, 256, 512, 1024\\
&Filter Size Layer 2 &(1$\times$1), (3$\times$3), (5$\times$5), (7$\times$7), (9$\times$9)\\
&Number of Filters Layer 3&16, 32, 64, 128, 256, 512, 1024\\
&Filter Size Layer 3&(1$\times$1), (3$\times$3), (5$\times$5), (7$\times$7), (9$\times$9)\\
&Number of Filters Layer 4&16, 32, 64, 128, 256, 512, 1024\\
&Filter Size Layer 4&(1$\times$1), (3$\times$3), (5$\times$5), (7$\times$7), (9$\times$9)\\

\hlineB{2}
\multirow{5}{*}{\textsc{ResNet}}&Number of Filters Stem &16, 32, 64, 128, 256, 512, 1024\\
&Number of Filters Projection Block &16, 32, 64, 128, 256, 512, 1024 \\
&Filter Size Projection Block &(1$\times$1), (3$\times$3), (5$\times$5), (7$\times$7), (9$\times$9)\\
&Number of Filters Bottleneck Block &16, 32, 64, 128, 256, 512, 1024\\
&Filter Size Bottleneck Block &(1$\times$1), (3$\times$3), (5$\times$5), (7$\times$7), (9$\times$9)\\

\hlineB{2}
\multirow{5}{*}{\textsc{SqueezeNet}}&Number of Filters Stem &16, 32, 64, 128, 256, 512, 1024 \\
&Number of Filters Stem &16, 32, 64, 128, 256, 512, 1024 \\
&Filter Size Fire Group One &(1$\times$1), (3$\times$3), (5$\times$5), (7$\times$7), (9$\times$9)\\
&Number of Filters Fire Group Two &16, 32, 64, 128, 256, 512, 1024\\
&Number of Filters Fire Block &16, 32, 64, 128, 256, 512, 1024 \\

\hlineB{2}
\multirow{5}{*}{\textsc{Bert}}&Dropout & 0.1, 0.3, 0.5, 0.7, 0.9\\
&Maximum Batch Size&6, 12, 16, 32, 64\\
&Maximum Sequence Length&13, 16, 32, 64, 128, 256\\
&Learning Rate &$1e^{-5}$, $2e^{-5}$, $3e^{-5}$, $4e^{-5}$, $5e^{-5}$\\
&Weight Decay &0, 0.1, 0.2, 0.3 \\
\hlineB{2}
\multirow{5}{*}{\textsc{Deepspeech}}&Num of epochs & 2, 4, 8, 16, 32\\
&Dropout & 0.1, 0.3, 0.5, 0.7, 0.9\\
&Maximum Batch Size& 16, 32, 64, 128, 256\\
&Maximum Sequence Length&16, 32, 64, 128, 256, 512, 1024\\
&Learning Rate &$1e^{-5}$, $2e^{-5}$, $3e^{-5}$, $4e^{-5}$, $5e^{-5}$\\

\doubleRule

\end{tabular}}
\label{tab:net-config}
\end{table}

\begin{table}[h]
\centering
\caption{OS-specific design options and their values.}
\resizebox{0.50\linewidth}{!}{
\begin{tabular}[t]{lll}
\doubleRule 
\textsc{Design Option} & \textsc{Value/Range}\\
\hline
Scheduler Policy&\texttt{CFP}, \texttt{NOOP} \\
Swappiness&10, 30, 60, 100\\
Dirty Background Ratio& 10, 50, 80\\
Dirty Ratio&5, 50\\
Cache Pressure & 100, 500\\
\doubleRule 
\end{tabular}}
\label{tab:os-config}
\end{table}

\begin{table}[h]
\centering
\caption{Hardware-specific design options and their values.}
\resizebox{0.50\linewidth}{!}{
\begin{tabular}[t]{llll}
\doubleRule 
 \multirow{2}{*}{\textsc{Design Option}}  & \multicolumn{2}{c}{\textsc{Value/Range}}\\
\cline{2-3}
&&Jetson Xavier\\

\hline
Num Active CPU& &1 - 6\\
CPU Frequency (GHz) & &0.3 - 2.3 \\
GPU Frequency (GHz)&&0.3 - 1.8 \\
EMC Frequency (GHz)&&0.3 - 2.0 \\

\doubleRule 
\end{tabular}}
\label{tab:hw-config}
\end{table}

\subsubsection{Objectives and Design Options}




We select two objectives: energy consumption and prediction error for
optimization for each architecture in our experiments. While we restrict
ourselves to two objectives, our methodology can be applied to an arbitrary
number of objectives. Depending on the particular hardware platform and DNN
architecture, we select 14-17 design options. Each platform and DNN has its own
specific hardware and DNN design options; OS-specific options are the same. We
consider 4 hardware-specific design options, 5 OS-specific options, and 5-8
DNN-specific options. Our chosen DNN-specific, OS-specific, and
hardware-specific design options are listed in Tables~\ref{tab:net-config}, \ref{tab:os-config}, and~\ref{tab:hw-config}, respectively. We
choose these options based on similar hardware's configuration guides/tutorials
and~\shortcite{halawa2017nvidia}. The choice of these design
options presents an interesting scenario for optimization based on how they
influence performance objectives because of the complex interactions of the
options. Hardware- and OS-specific options like the number of active CPUs or the scheduler policy affect only energy consumption, whereas  DNN options like filter size or the number of filters affect both energy consumption and prediction error. Depending on the DNN architecture, we use either Keras (Tensorflow as backend) or PyTorch as the compiler for training and prediction (see Table \ref{tab:datasets} for
details).




\begin{table}[h]
\centering
\caption{The list of hyperparameters used for the surrogate models.}
\resizebox{0.6\linewidth}{!}{
\begin{tabular}[t]{lll}
\doubleRule
\textsc{Surrogate} & \textsc{Hyperparameters} &\textsc{Value}\\
\hline
 \multicolumn{1}{l}{\multirow{3}{*}{\rotatebox{0}{\textsc{GP}}}}&{Kernel}&Squared Exponential \\
&Num Restarts&20\\
&Optimizer& L-BFGS-B\\
&$\alpha$&$1e^{-10}$\\
\hline
 \multicolumn{1}{l}{\multirow{3}{*}{\rotatebox{0}{\textsc{RF}}}}&Num Trees&128 \\
&Min Split Variable& 2\\
&Min Impurity Split&$1e^{-7}$\\
\doubleRule
\end{tabular}}
\label{tab:hyperparam-surrogate}
\end{table}

\subsubsection{Setting}
To initialize \tool, we measure the prediction error and energy
consumption of $20$ randomly selected designs from the design space of a
particular DNN system. As energy consumption measurements tend to be noisy,
we take $10$ repeated measurements for a particular design $\bm{x}$ and consider
the median. We do not repeat prediction error measurements as they are not noisy. We use two different
surrogate models, Gaussian process (GP) and Random Forest (RF), termed \textsc{FlexiBO-GP} and \textsc{FlexiBO-RF}, respectively. Details of the hyperparameters used for both GP and RF are provided in Table \ref{tab:hyperparam-surrogate}. We use the Wall-Clock Time $t_{wc,i}$ required to evaluate an objective $f_i$ as the computational effort
and run experiments with three different objective evaluation cost functions: (i) Logarithmic cost (LC): $\theta_i=\log\left(t_{wc,i} \right)$, (ii) Ratio cost (RC): $\theta_i = \frac{t_{wc,i}}{\min (t_{wc,i})_{1 \leq i \leq n}}$
and, (iii) Constant cost (CC):
$\theta_i=1$ to simulate a method that is not cost-aware. At each iteration, \tool \ recommends a design and an objective for evaluation. Depending on the objective selected for evaluation, we take the following actions.

\begin{itemize}
    \item[$\blacktriangleright$] The recommended objective is \emph{prediction error}.
    \begin{itemize}
    \item[$\checkmark$] We retrain a DNN with the DNN-specific options
of the selected design if no pre-trained model for the DNN-specific options
exists. We reuse the pre-trained model otherwise. We measure  prediction error.
\end{itemize}
\end{itemize}

\begin{itemize}
    \item[$\blacktriangleright$] The recommended objective is \emph{energy consumption}.
    \begin{itemize}
    \item[$\checkmark$] If no pre-trained model with the
DNN-specific options of the selected design exists, we use a model whose size is
the same as that of the model obtained after random initialization of the
weights using the DNN-specific design options, else we reuse the pre-trained
model. We measure the energy consumption for the selected design.
\end{itemize}
\end{itemize}
Figure \ref{fig:implementation} gives a high-level overview of our experimental
setup.
We implement \tool \ in a distributed manner where the training of a DNN is done remotely on virtual machine instances with $8$ NVIDIA Tesla K80 GPU deployed on the Google cloud and the measurements and optimization algorithms run locally on resource-constrained Jetson devices, i.e.\ Xavier. Our experiments took a total of 5552.4 hours of wall-clock time to complete. 

\paragraph{Hypervolume Error.}
The Pareto hypervolume ($\mathfrak{hv}$) is commonly used to measure the quality of an estimated Pareto front $\mathfrak{\hat{F}^*}$\shortcite{cao2015using,Zitzler1999HV}.
As seen in Equation~\ref{eq_hypervolume}, it is defined as the volume enclosed by the estimated Pareto front $\mathfrak{\hat{F}^*}$ and a user-defined reference point $r$ in the objective space, in our case the origin of the coordinate system.
\begin{equation}
    \label{eq_hypervolume}
    \mathfrak{hv}(\mathfrak{\hat{F}^*}, r) = V(\cup_{s \in \mathfrak{\hat{F}^*}}\{q | r \preceq q \preceq s\})
\end{equation}
The hypervolume error $\eta$ is defined as the difference between the hypervolumes of the \emph{true Pareto front} $\mathfrak{F}^*$ and the estimated Pareto front $\hat{\mathfrak{F}}^*$. 


\begin{equation}
    \eta = \mathfrak{hv}(\mathfrak{F^*}, r) - \mathfrak{hv}(\mathfrak{\hat{F}^*}, r)
\end{equation}
\begin{figure}[t]
    \centering
     \includegraphics[width=0.32\textwidth]{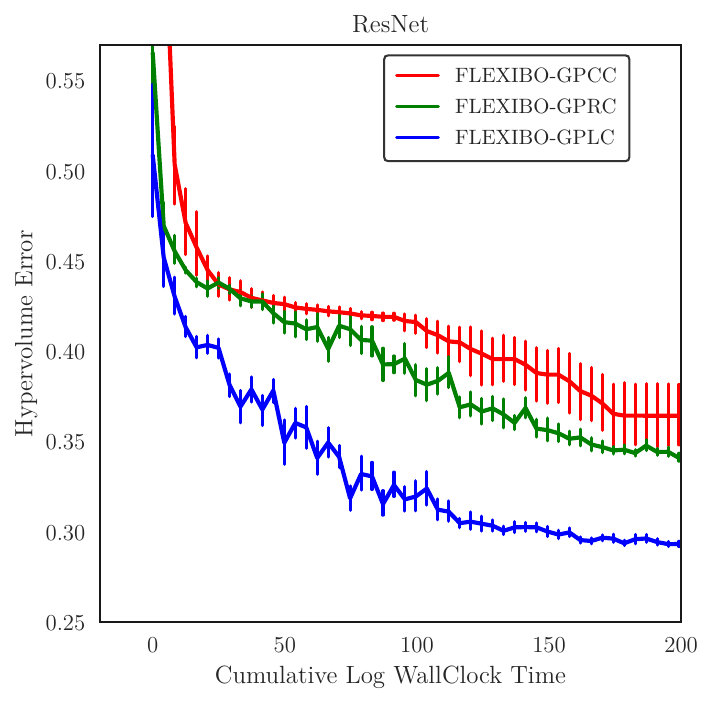}  
     \includegraphics[width=0.32\textwidth]{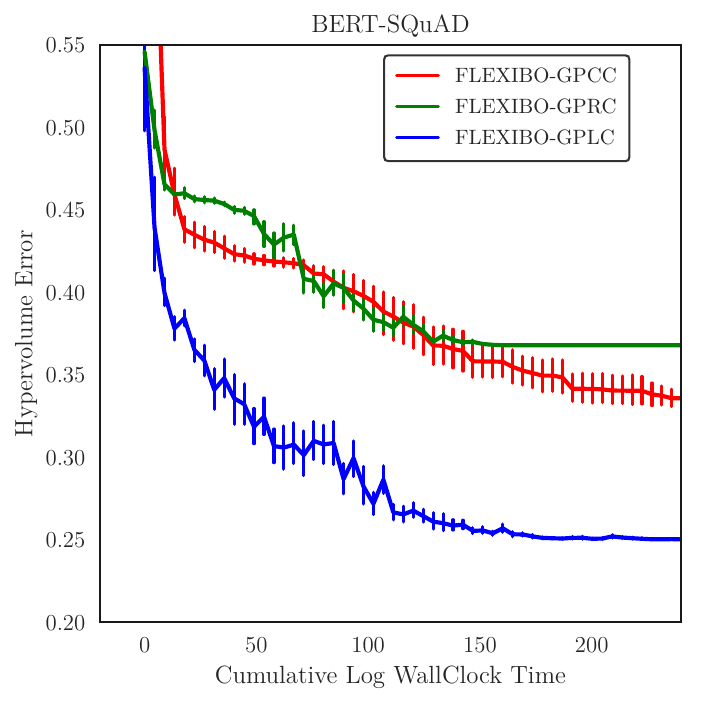} 
     \includegraphics[width=0.32\textwidth]{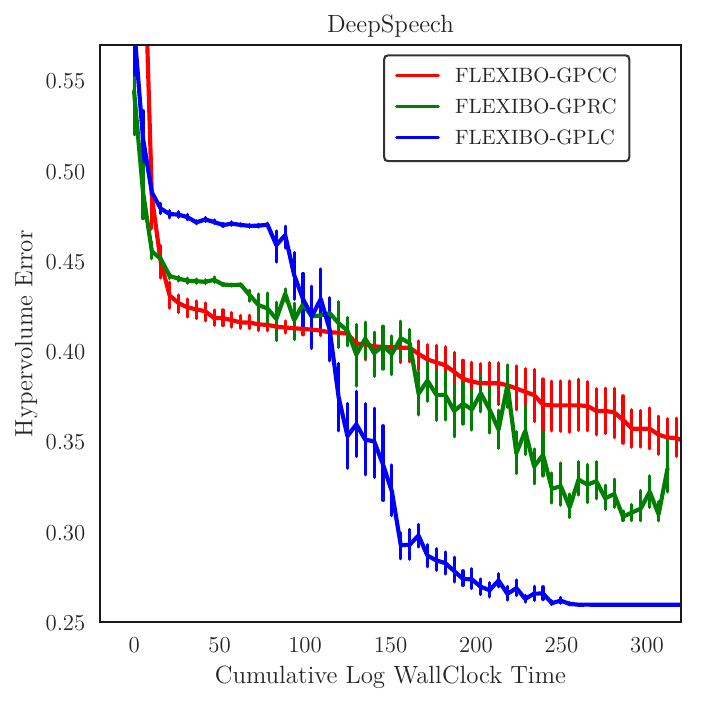} 

    \caption{Comparison of hypervolume error obtained by \tool \ using different cost functions.}
    \label{fig:hv-cost-function}
    
\end{figure}


\begin{figure}[t]
    \centering
     \includegraphics[width=0.32\textwidth]{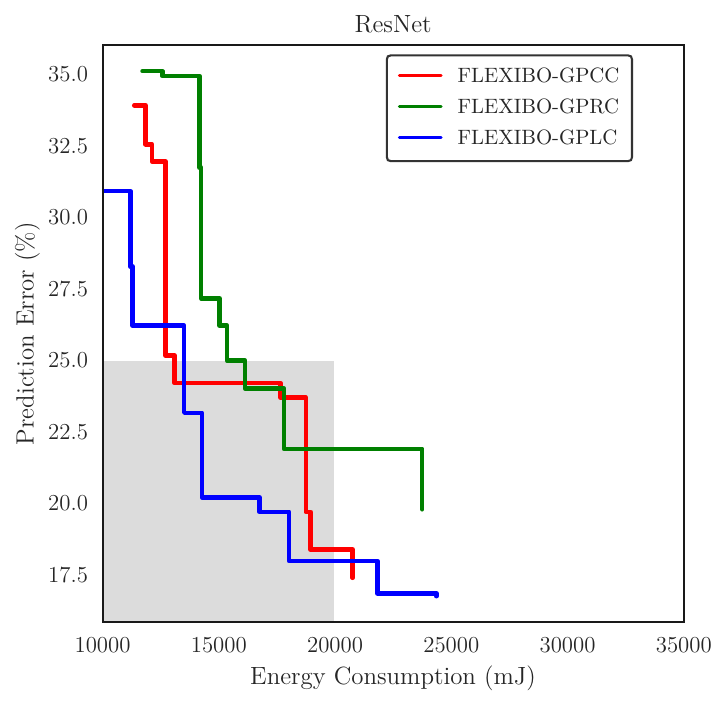}  
     \includegraphics[width=0.32\textwidth]{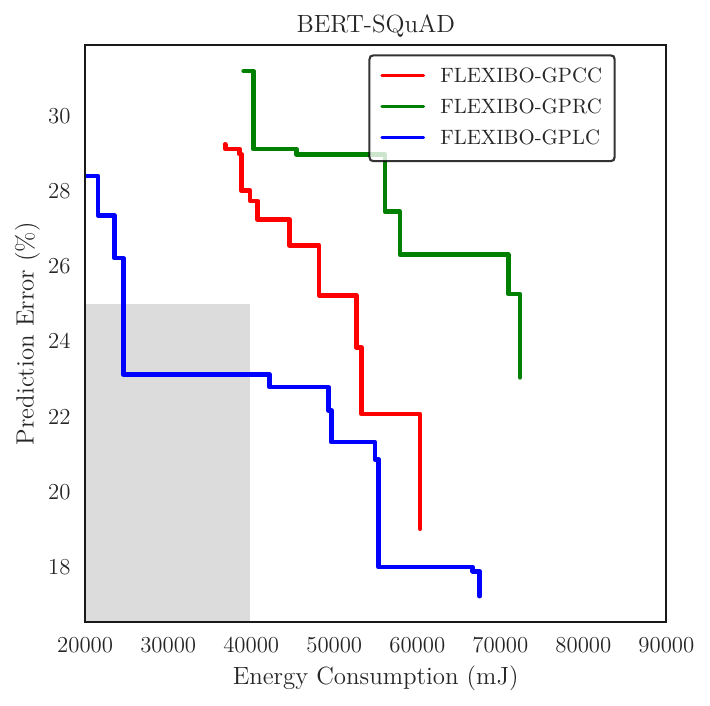}  
     \includegraphics[width=0.32\textwidth]{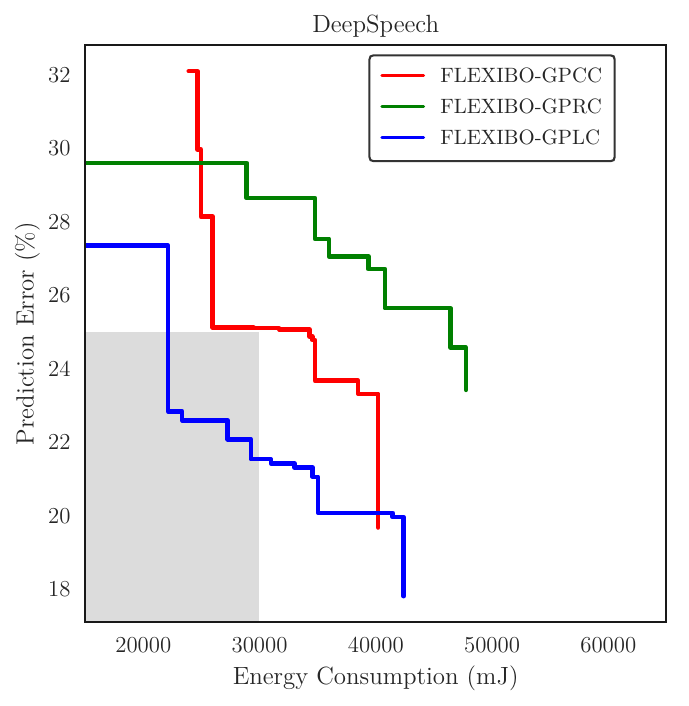}  

    \caption{Comparison of Pareto optimal designs obtained by \tool \ using different cost functions.}
    \label{fig:pf-cost-function}
    
\end{figure}
\begin{figure}[h]
    \centering
    \includegraphics[width=0.415\textwidth]{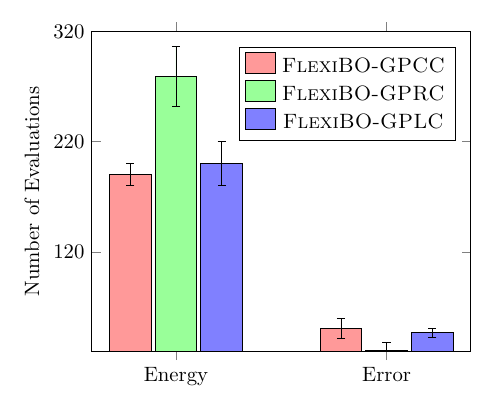}  
    \includegraphics[width=0.4\textwidth]{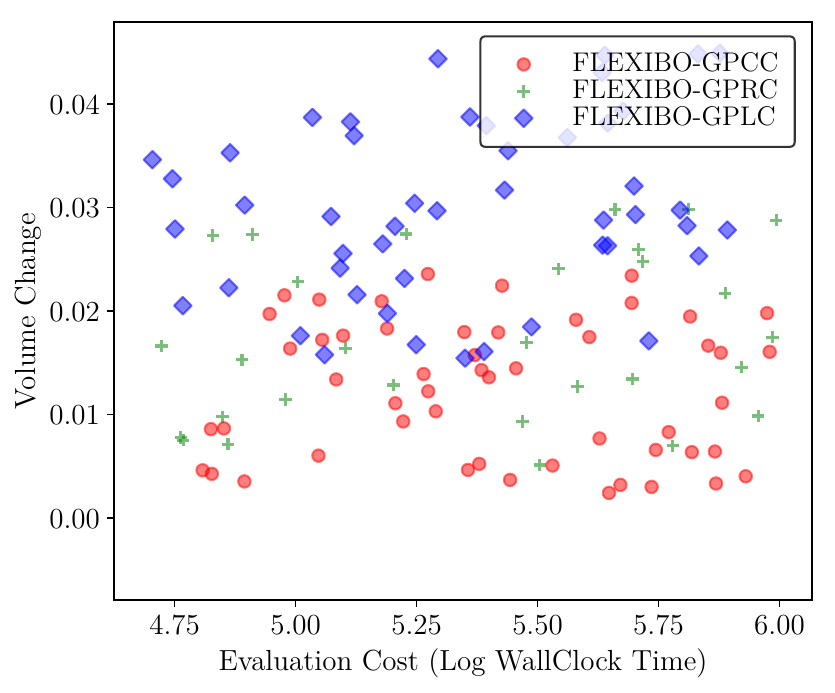}  
    \caption{ (a) Number of evaluations across each objective by \tool \ using different cost functions (b) \gplc \ achieved higher change of volume of the Pareto region with the recommended design and objective when compared to others.}
    \label{fig:analysis-cost-function}
    
\end{figure}

We evaluate the quality of the obtained Pareto fronts using the hypervolume error and the cumulative log wall-clock time as the objective evaluation cost required to obtain it.
As the actual Pareto fronts are unknown, we approximate them by combining the Pareto fronts obtained by the different optimization methods considered in our experiments.

\subsection{Experimental Results}
Given the same wall-clock time,
we observe the hypervolume error obtained by Pareto fronts identified by the
different optimization methods.
Furthermore, to assess the quality of the Pareto fronts, we compare the number
of designs in the target region of the objective space. Our target region is the
region where the prediction error is less than 25\% and energy consumption is
less than the first quartile. Note that energy consumption is specific to the hardware platform. 

\begin{figure}[t]
    \centering
     \includegraphics[width=0.32\textwidth]{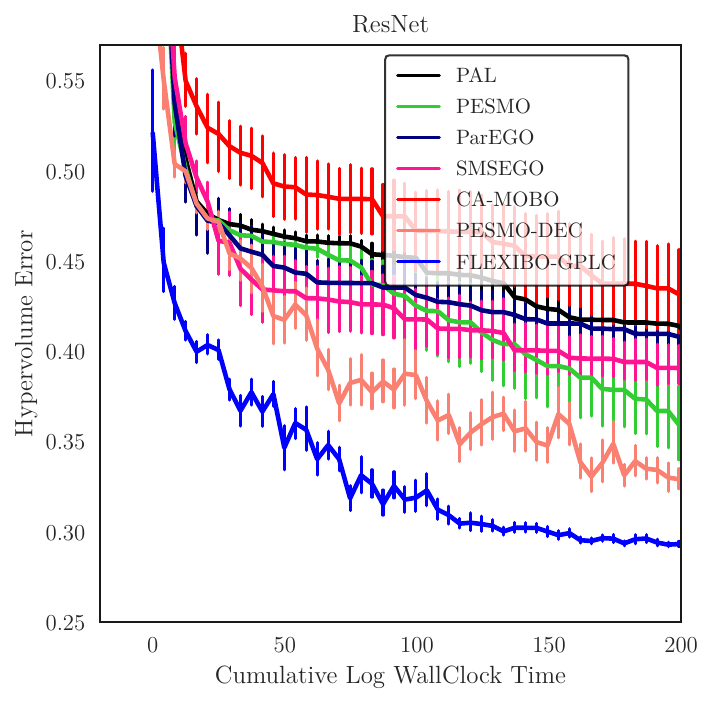} 
     \includegraphics[width=0.32\textwidth]{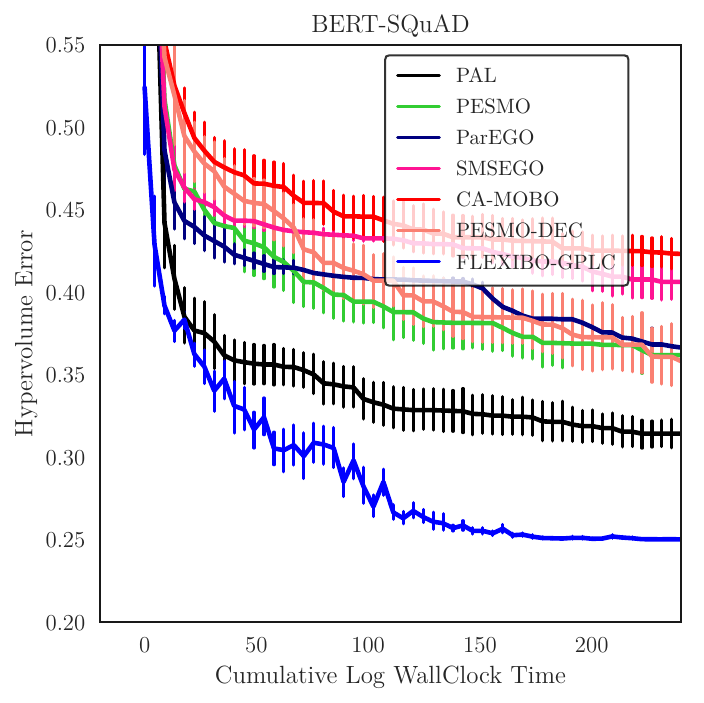} 
     \includegraphics[width=0.32\textwidth]{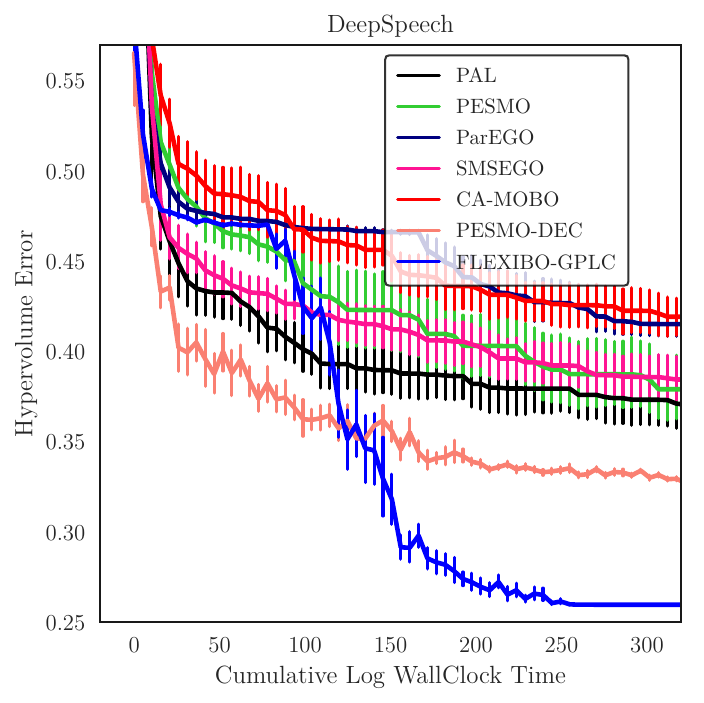} 
   \caption{Comparison of hypervolume error obtained by \tool \ and other approaches for DNNs for object detection, NLP, and speech recognition applications.}
    \label{fig:hv-dnn-app}
\end{figure}

\begin{figure}[t]
    \centering
     \includegraphics[width=0.32\textwidth]{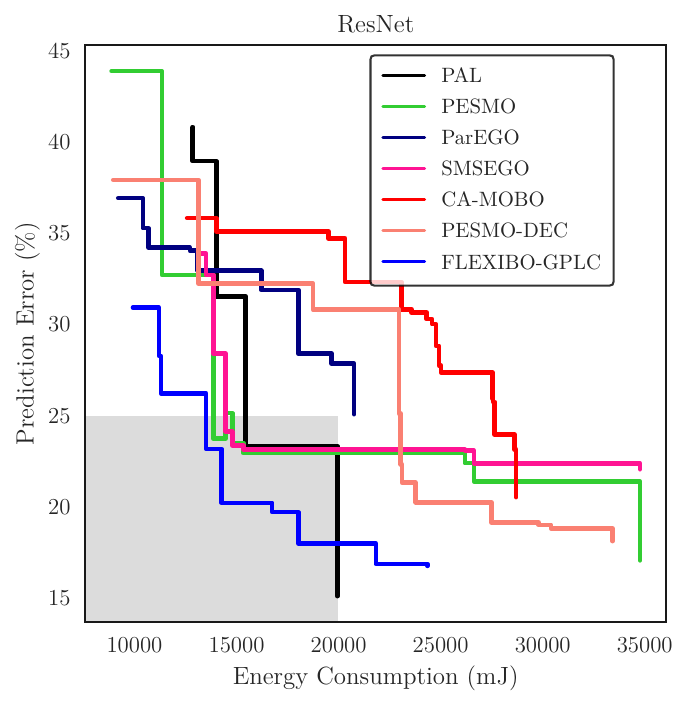} 
     \includegraphics[width=0.32\textwidth]{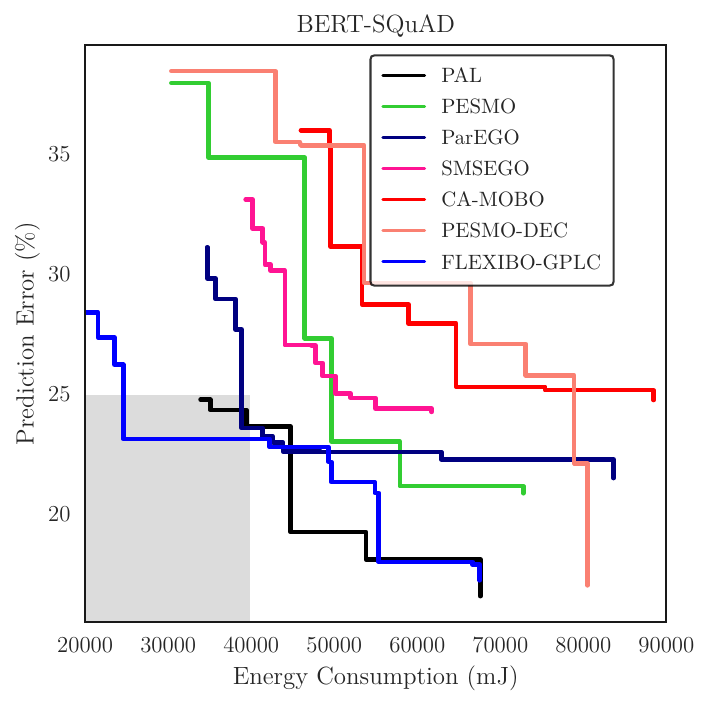} 
     \includegraphics[width=0.32\textwidth]{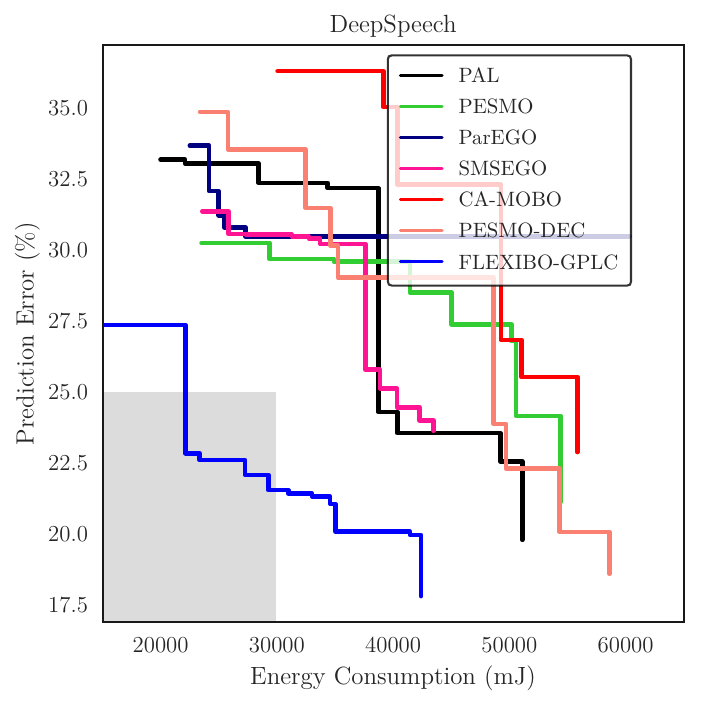} 
     \caption{Comparison of Pareto fronts obtained by \tool \ and other approaches for DNNs for object detection, NLP, and speech recognition applications.}
    \label{fig:pf-dnn-app}
\end{figure}
\subsubsection{RQ1: Determination of objective evaluation cost function}
Figures~\ref{fig:hv-cost-function} and~\ref{fig:pf-cost-function} show the results for optimizing prediction error and energy consumption with different cost functions. \gplc \ has lower hypervolume error (shown in Figure~\ref{fig:hv-cost-function}) and a higher number of designs in the target region (shown in Figure~\ref{fig:pf-cost-function}) than \gprc \ and \gpcc.  To better understand the effect of different cost functions, we also look at the behavior of \tool \ in Figure~\ref{fig:analysis-cost-function}(a) and \ref{fig:analysis-cost-function}(b). Cost-unaware \gpcc \ greedily selects the design and objective across which the volume change is maximal for evaluation. As a result, \gpcc \ wastes resources by selecting expensive evaluations for little gain. This is evident from \ref{fig:analysis-cost-function}(b) which indicates that the reduction of volume by the designs selected by \gpcc are lower considering the evaluation cost. A larger change in the volume of Pareto region is desired as it indicates higher information gain. Objective evaluation cost function in \gprc \ is skewed towards selecting the objective with lower evaluation cost and evaluates a higher number of designs for the less expensive objective e.g., energy consumption (shown in Figure~\ref{fig:analysis-cost-function}(a)). However, it achieves lower change of the volume of the Pareto region than \gplc. \gplc \ on the other hand selects designs that achieved a larger volume change across the Pareto region (Figure~\ref{fig:analysis-cost-function}(b)) and therefore a better choice than others. 


\subsubsection{RQ2: Effectiveness of \tool}
\noindent \paragraph{Effectiveness across DNNs of different applications.}

Figures~\ref{fig:hv-dnn-app} and~\ref{fig:pf-dnn-app} show the effectiveness analysis of \tool \ across different applications. In Figure~\ref{fig:hv-dnn-app} we observe that \gplc \ outperforms other methods in finding Pareto fronts with lower hypervolume error for each of the applications. For example, \tool \ achieves $22.4\%$ lower hypervolume error than CA-MOBO in \textsc{DeepSpeech}. In Figure~\ref{fig:pf-dnn-app}, we observe that \gplc \ is able to find a higher number of designs in the target region than other methods for \textsc{ResNet}, \textsc{BERT-SQuAD}, and \textsc{DeepSpeech}. 
\begin{figure}[h]
    \centering
     \includegraphics[width=0.24\textwidth]{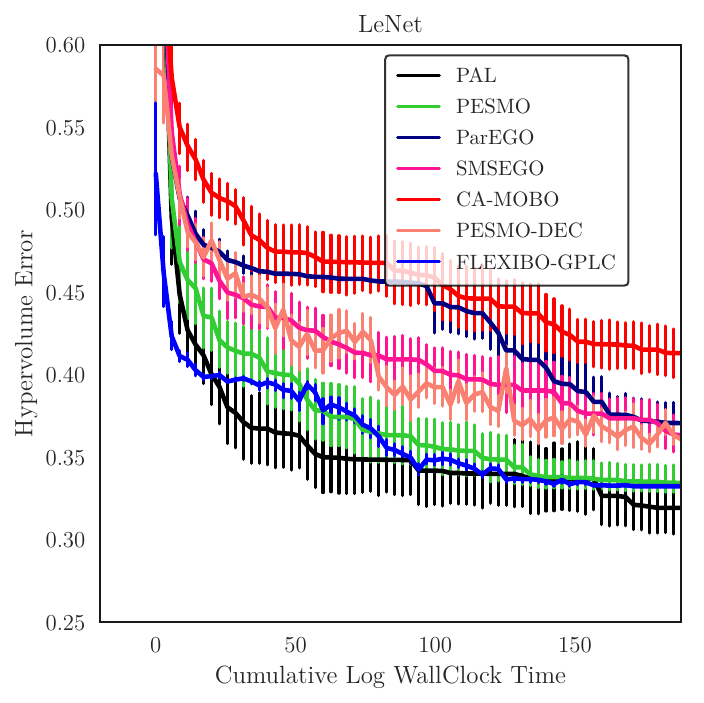} 
     \includegraphics[width=0.24\textwidth]{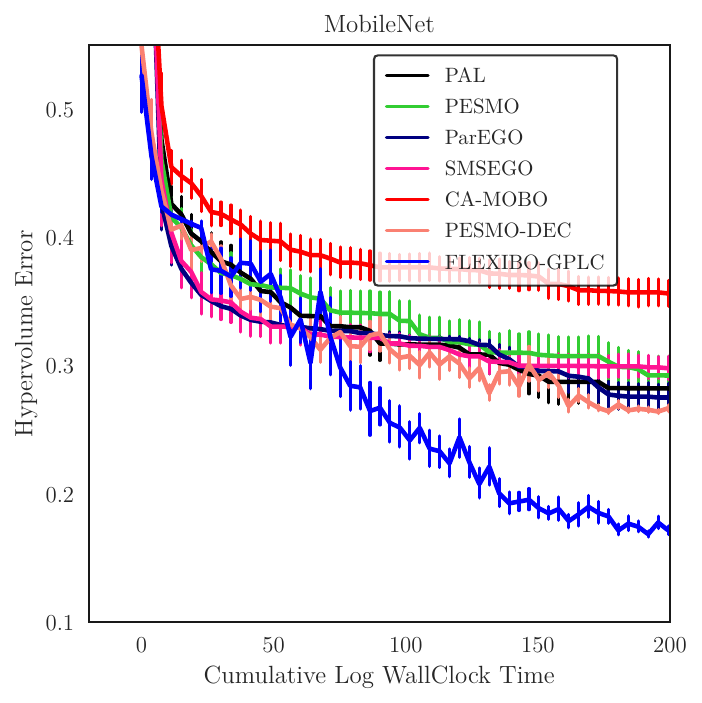} 
     \includegraphics[width=0.24\textwidth]{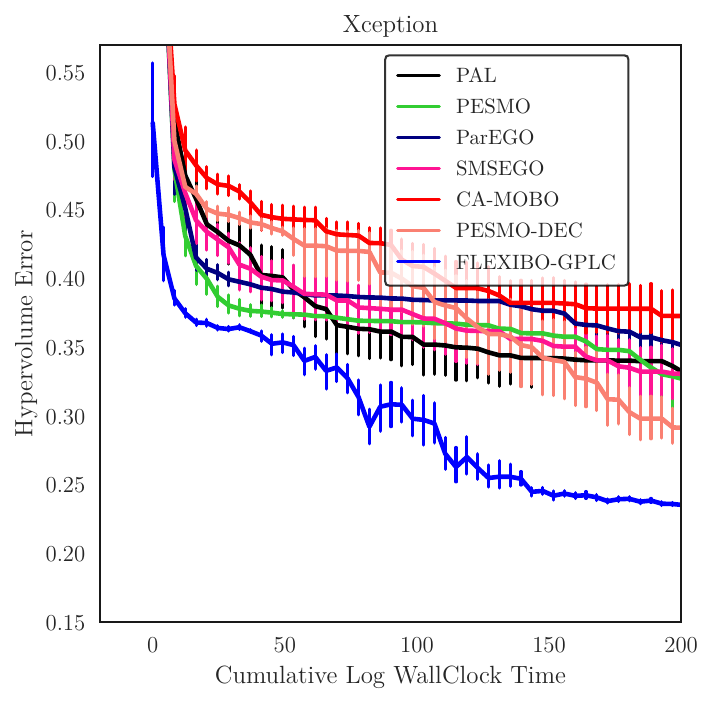} 
      \includegraphics[width=0.24\textwidth]{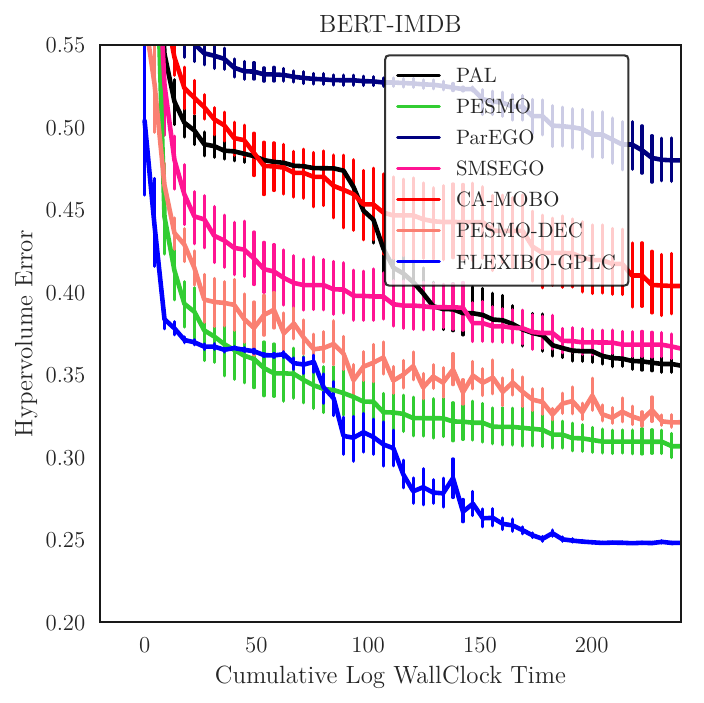} 
    \caption{Comparison of hypervolume error obtained by \tool \ and other approaches for DNNs of different sizes.}
    \label{fig:hv-dnn-size}
\end{figure}

\begin{figure}[h]
    \centering
     \includegraphics[width=0.24\textwidth]{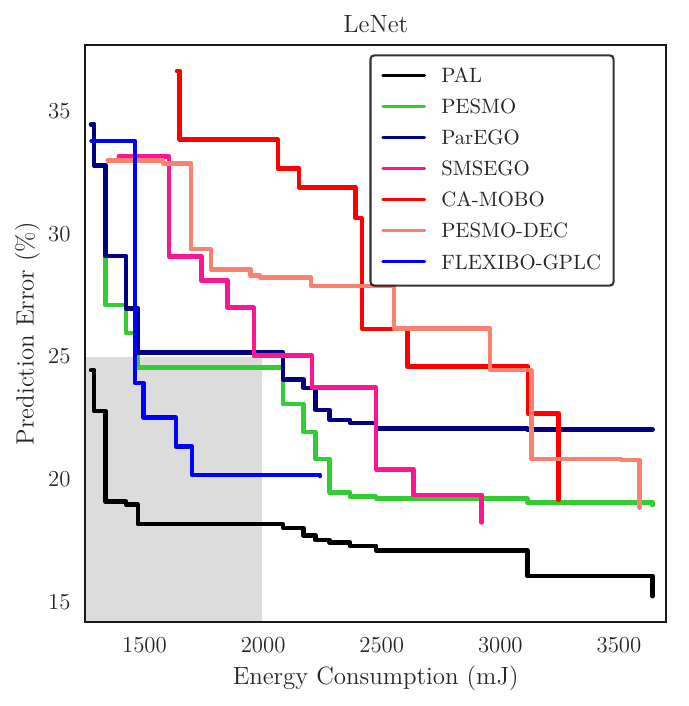} 
     \includegraphics[width=0.24\textwidth]{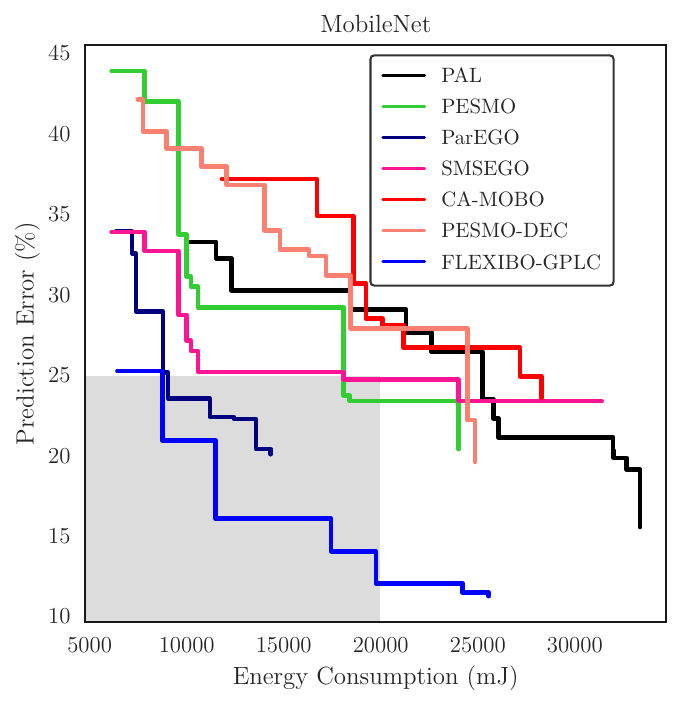} 
     \includegraphics[width=0.24\textwidth]{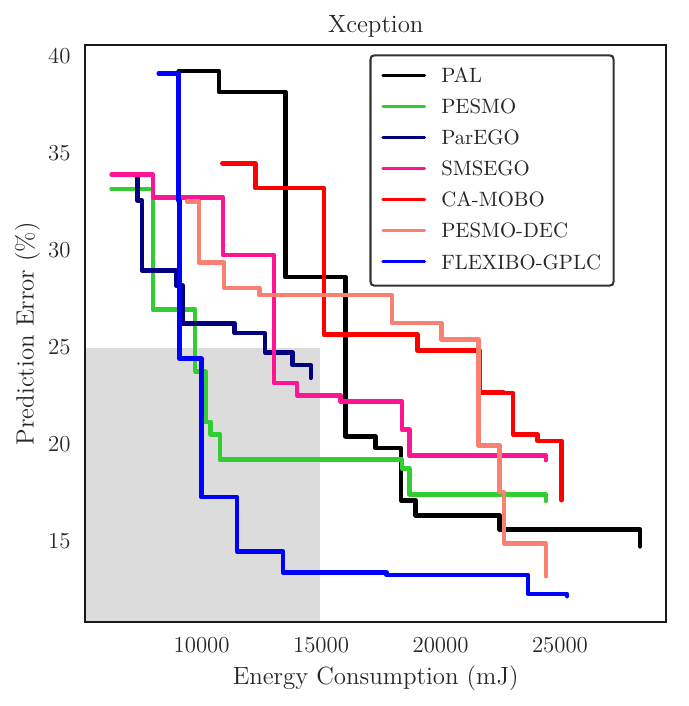} 
      \includegraphics[width=0.24\textwidth]{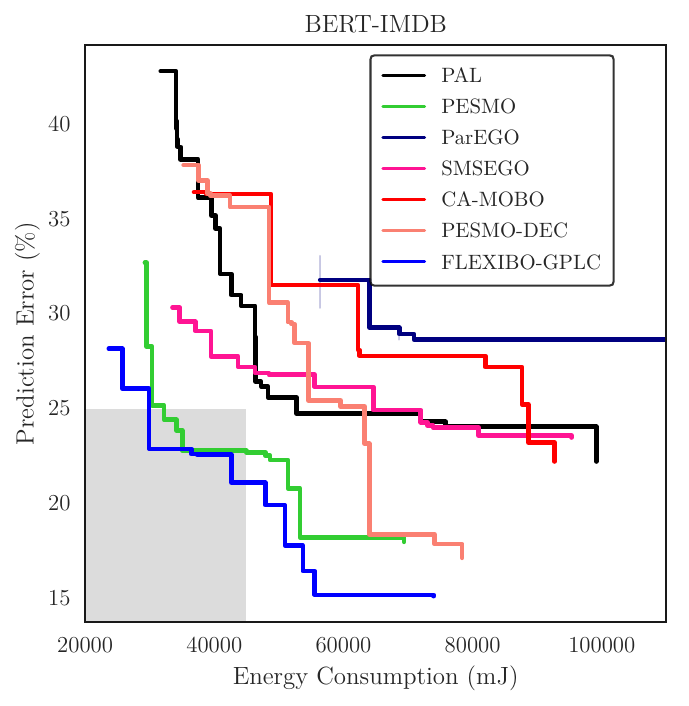} 
    \caption{Comparison of Pareto fronts obtained by \tool \ and other approaches for DNNs of different sizes.}
    \label{fig:pf-dnn-size}
\end{figure}

\begin{figure}[h]
    \centering
     \includegraphics[width=\textwidth]{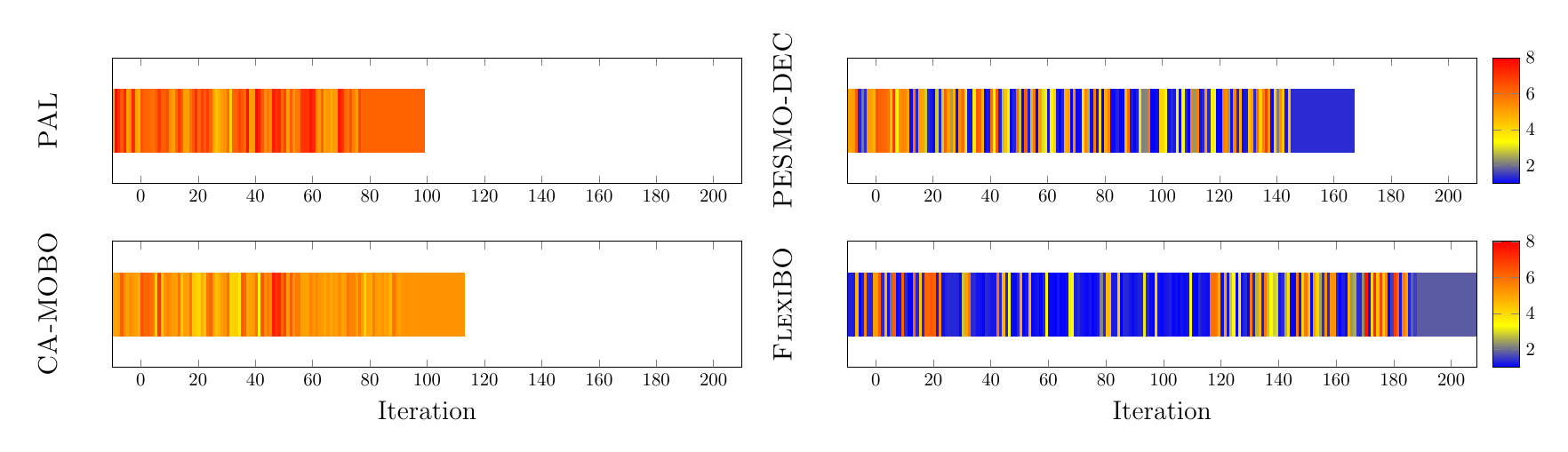} 
 
    \caption{\tool \ utilizes resources more efficiently than other approaches when the difference of evaluation cost between objectives is high. The colors indicate the evaluation cost. \tool is able to run for more iterations as it expends less of the evaluation budget at each one.}
    \label{fig:resource-utilization}
\end{figure}

\noindent \paragraph{Effectiveness across DNNs of different sizes.}

Figures~\ref{fig:hv-dnn-size} and~\ref{fig:pf-dnn-size} show the effectiveness analysis of \tool across different-size DNNs. We make the following observations:
(a) as shown in Figure \ref{fig:hv-dnn-size}, we find that \gplc \ outperforms other
methods in finding Pareto fronts with lower hypervolume error across all applications (e.g.\ $22.3\%$ lower
for IMDB than CA-MOBO), (b) in Figure~\ref{fig:pf-dnn-size}, we observe that \gplc \ is able to find a higher number of designs in the target region than other methods for Xception, MobileNet, and BERT-IMDB. For LeNet, \pal \ achieves a $3.6\%$ lower hypervolume error than \gplc. LeNet is a small architecture and \tool \ performs poorly for such small architectures as the effect of selecting designs based on change of volume of the Pareto region per cost is less pronounced than for larger architectures (Xception or BERT-IMDB etc).


\begin{table}[h]
  \caption{Time (in seconds) required for one iteration with and without objective evaluation across all architectures.}
  \resizebox{\textwidth}{!}{%
  \begin{tabular}{lccccccc}
    \toprule
    
    & \tool \ & \pesmo \ & \pesmodec \ & \pal \ & \camobo \ &\parego \ & \smsego \ \\
    \hline
    \midrule
    \multicolumn{1}{l}{\multirow{1}{*}{\rotatebox{0}{\textsc{No Evaluation}}}}&79.9$\pm$7.4&64.8$\pm$5.1&66.9$\pm$ 5.4&178.2$\pm$ 10.6&71.2$\pm$ 8.4&56.2&48.4$\pm$4.3\\
    \hline
    \multicolumn{1}{l}{\multirow{1}{*}{\rotatebox{0}{\textsc{With Evaluation}}}}&1417.2$\pm$ 97.6&9133.7$\pm$448.8&4400.4$\pm$225.9&8764.3$\pm$469.5&8411.5$\pm$365.4&8656.9$\pm$220.3&9020.9$\pm$306.3\\
    \bottomrule
  \end{tabular}}
  \label{tab:time-acq}
\end{table}

We observe that approaches other than \tool \ cannot make the best use of the allocated budget as they evaluate the more expensive objectives more than the cheap objectives. As the expensive objectives can be selected any time (even for little gain), this strategy is wasteful when limited resources are available. \tool \ makes better use of the resources by evaluating the cheaper objectives more in the earlier iterations and thus gaining a better understanding of the design space and only later evaluating the costly objective (Figure~\ref{fig:resource-utilization}). We also find that \tool \ is able to evaluate more designs by prudently selecting the objectives across which to evaluate it (Figures~\ref{fig:hv-dnn-size} and~\ref{fig:pf-dnn-size}). A cost-aware decoupled approach is clearly useful for scenarios where the evaluation budget is limited.



\noindent \paragraph{Comparison of average time required for modeling.}
Table~\ref{tab:time-acq} shows the average time required for one iteration for different multi-objective optimization methods, averaged across all architectures. Though \tool \ requires more time to compute the acquisition function (no evaluation) than others, the time required for one iteration including the objective evaluation time in \tool \ is $5.6 \times$ lower than the next best method \parego. 

\subsubsection{RQ3: Sensitivity Analysis}

\begin{figure}[h]
    \centering
     \includegraphics[width=0.32\textwidth]{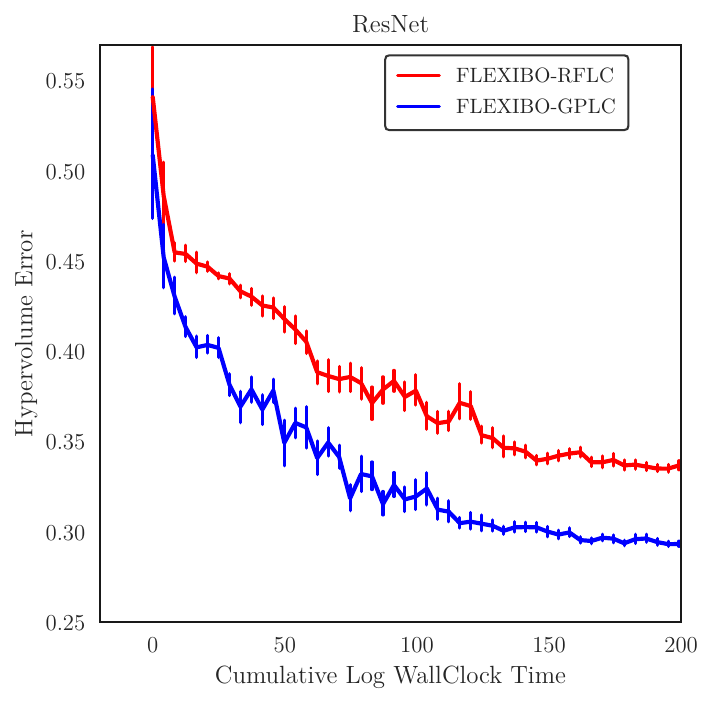}  
      \includegraphics[width=0.32\textwidth]{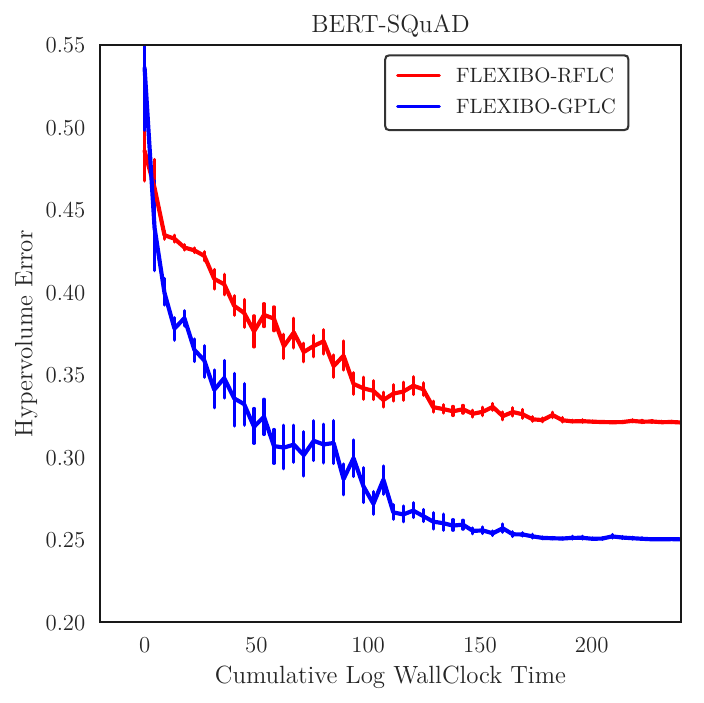} 
     \includegraphics[width=0.32\textwidth]{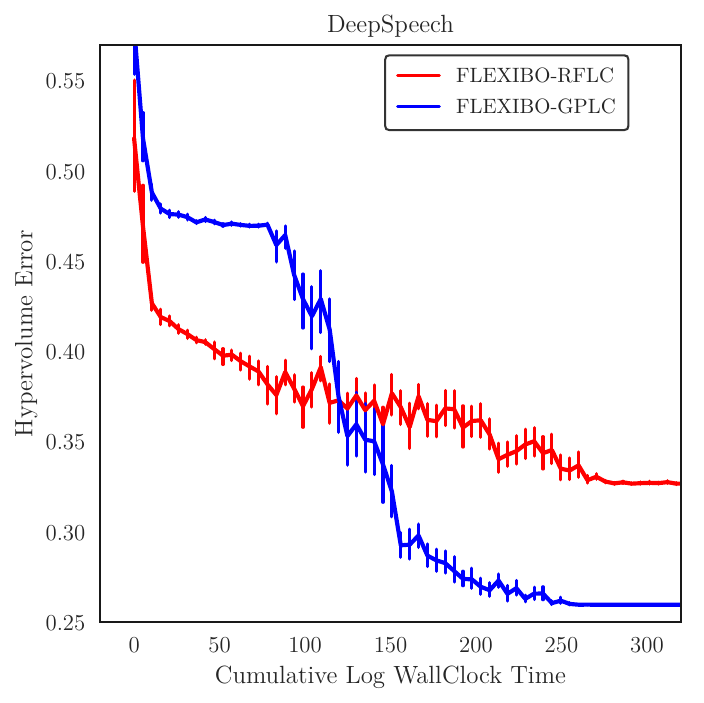} 

    \caption{Comparison of hypervolume error obtained by \tool \ with different surrogate models.}
    \label{fig:hv-surrogate}
    
\end{figure}


\begin{figure}[h]
    \centering
     \includegraphics[width=0.32\textwidth]{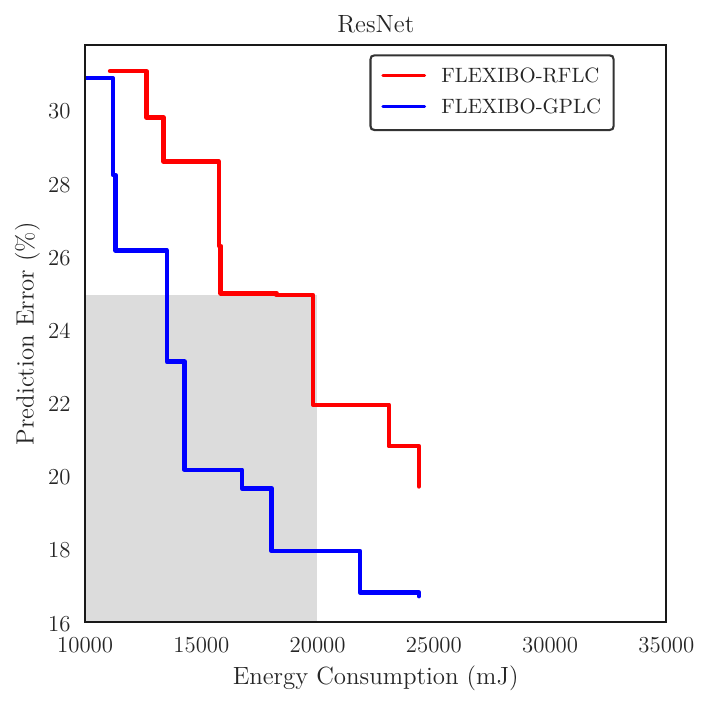}  
     \includegraphics[width=0.32\textwidth]{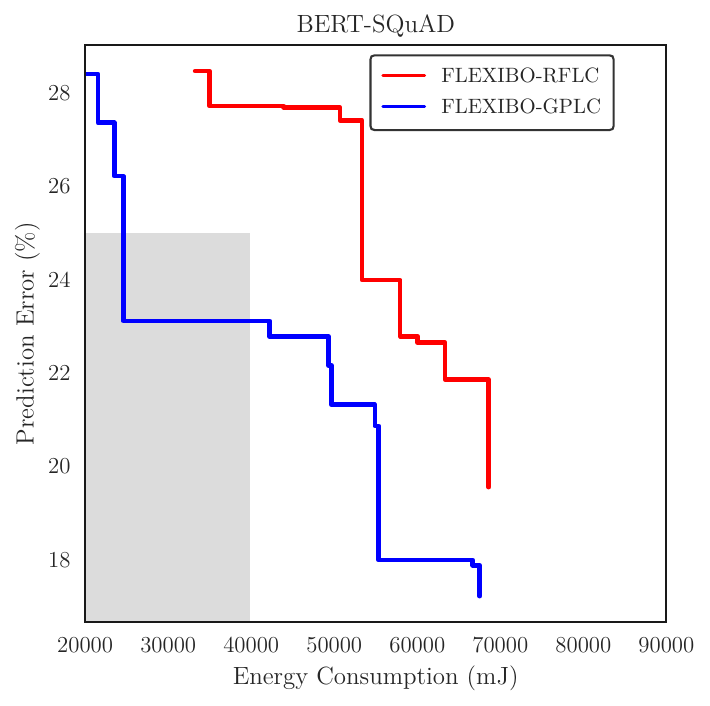}   
     \includegraphics[width=0.32\textwidth]{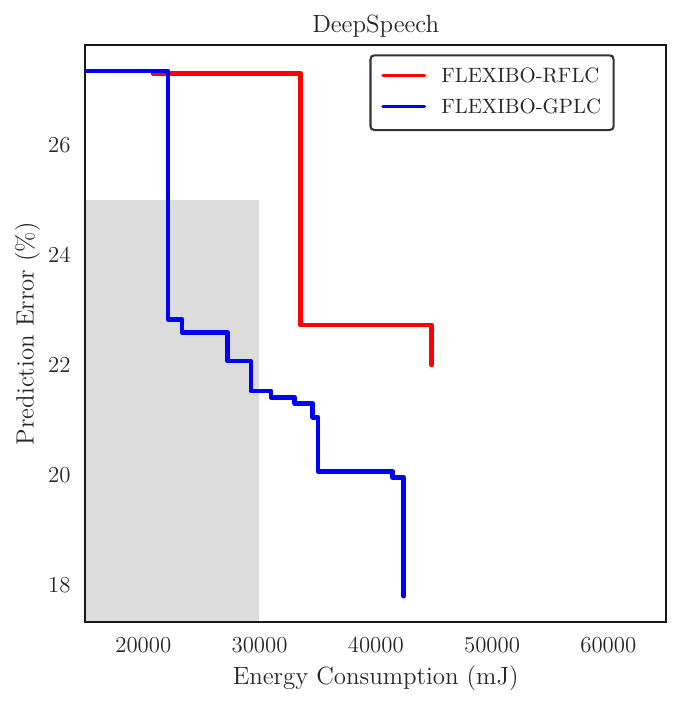}    

     \caption{Comparison of Pareto fronts obtained by \tool \ with different surrogate models.}
    \label{fig:pf-surrogate}
    
\end{figure}

        
    

\noindent \paragraph{Different surrogate models.}
We compare the performance of the different variants of
\tool: \gplc \ and \rflc \ that use
log objective evaluation cost. We used this cost function as it is a better choice than ratio and constant cost functions. Figures~\ref{fig:hv-surrogate} and~\ref{fig:pf-surrogate} show the hypervolume error and quality of the obtained Pareto fronts. We find that across all architectures \gplc \ outperforms \rflc \ (\gplc \ has lower hypervolume error and higher number of designs in the target region). 

    




\section{Conclusion}

In this work, we proposed a novel cost-aware acquisition function for Bayesian multi-objective
optimization called \tool. Instead of evaluating all objective functions, \tool \ automatically chooses the one that provides the highest benefit,
weighted by the cost to perform the evaluation. We showed the promise of our
approach through an extensive and thorough evaluation of seven different DNN
architectures over a large design space on resource-constrained hardware platforms.
Our experimental results show that \tool \ performs better than current
state-of-the-art approaches in most cases, both in terms of the quality of the
obtained Pareto fronts and the cost necessary to obtain them.

\section*{Acknowledgements} This work has been supported, in part, by NSF (Awards 2007202 and 2107463), as well as Google and Chameleon Cloud (provided cloud resources for the experiments). We are grateful to all who provided feedback on the earlier versions of this work, including Luigi Nardi (and several members of his group at Lund), Mohammad Ali Javidian, Md Abir Hossen, several members of 
ABLE Research Group at CMU, and anonymous reviewers of AutoML'22. 

\bibliography{FlexiBO}
\bibliographystyle{theapa}
\newpage
\section{Appendix}
\label{sec:theory}
In this section, we analyze the sample complexity of \tool.  
\begin{table}[t]
\caption{List of symbols and their descriptions.}
\resizebox{0.6\linewidth}{!}{
\begin{tabular}{lp{0.5\textwidth}}
\toprule
\textbf{Symbol} & \textbf{Description}\\
&\\
\hline
$t$ &  Number of iteration \\ $P_R $ &  Pareto region\\
$T$ &  Total number of iterations \\ $\mathfrak{F}_{pess} $ &  Pessimistic Pareto front\\
$n$ &  Number of objectives \\ $V_t$ &  Volume of $P_{R}$ at iteration $t$\\
$\mu$ &   Posterior mean \\ $\sigma $ &  Posterior standard deviation\\
$\mathfrak{U}$ &  Non-dominated points set \\ $\theta_{t,i}$ &  Evaluation cost of an objective $f_i$\\
$\bm{x}$ & A design \\ $r$ &  A reference point\\
$\mathfrak{X}$  & Design space \\ $\Delta V_t $ &  Change of volume of the $P_R$ at iteration $t$\\
$f$ &  An objective \\ $R_t(\bm{x})$ &  Uncertainty region of a point $\bm{x}$ at iteration $t$\\
$\beta_t$ & Scaling parameter value at iteration $t$ \\ $\mathfrak{M}$& Surrogate model\\
$S_i$  & Evaluated points set for objective $f_i$ \\ $\mathfrak{F}^*$ &  Optimal Pareto front\\
$\mathfrak{X}^*$ & Pareto-optimal set \\ $N_0$ & Number of initial samples \\
$\eta$ & Pareto hypervolume error \\ $\mathfrak{hv}$ & Pareto hypervolume \\
$\theta_{T}$ & Total objective evaluation cost \\ $\delta$ & Probability \\
$\gamma_T$ & Maximum information gain \\ $\alpha$ & Acquisition function  \\
$\Delta^n$ & n-simplex  \\ $V(\Delta^n)$ & Volume of an n-simplex\\
$k$ & Co-variance  \\ $y_i$ & Actual value of an objective $f_i$  \\
$v$  & Measurement noise \\ $\hat{\mathfrak{F}^*}$ &  Approximate optimal Pareto front\\
\bottomrule
\end{tabular}}

\label{tab:notations}
\end{table}
Let us assume that the maximum iteration within budget $\theta_T$ is $T$. By extending the theory from \pal, we derive the convergence rate of our proposed \tool \ algorithm.  ~\cite{zuluaga2013active} demonstrated that the critical quantity governing the convergence rate is given by the following:

\begin{equation}
\gamma_T =\max_{\bm{y}_1...\bm{y}_T} I(\bm{y}_1...\bm{y}_T;\bm{f}),    
\end{equation}
i.e., maximum reduction of uncertainty achievable by sampling $T$ designs. For \tool, the maximum reduction of uncertainty corresponds to the maximum change of the volume of the Pareto region $\Delta V$ and the above equation can be written as:
\begin{equation}
\gamma_T =\max_{\bm{y}_1...\bm{y}_T} \Delta V(\bm{y}_1...\bm{y}_T;\bm{f}),    
\end{equation}
Similar to \shortcite{srinivas2012information,zuluaga2013active}, we also establish $\gamma_T$ as the key quantity in bounding the hypervolume error $\eta$ in our analysis. The following theorem is our main theoretical result. 




\begin{figure}
    \centering
    \includegraphics{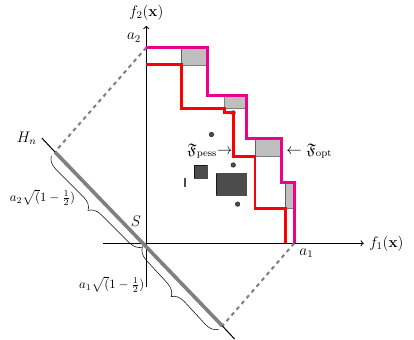}
    \caption{Example of hypervolume error bound for two objectives.}
    \label{fig:lem7}
\end{figure}

\begin{theorem}
\label{thm:1}
Let $\delta \in (0,1)$. \tool \ running with $\beta_t= 2/9 \log(n|\mathfrak{X}|\pi^2t^2/{6\delta})$ would achieve a maximum hypervolume error
of $\eta$ of the Pareto front obtained inside total cost $\theta_T$ with probability $1-\delta$. 

\begin{equation}
    \eta \leq \frac {\sqrt{n}a^{n-1}} {(n-1)!} \left\{ a^n - 2\Big(\frac{C_1 \beta_T \gamma_T}{2}\Big)^{n/2}{\frac{(\frac{\theta_T}{\theta_x}(n+1))^{1/2}}{n!}}  \right\} 
\end{equation}
Here, $a_i=\max_{\bm{x} \in \mathfrak{X_m}, 1 \leq i \leq n} {\sigma_i (\bm{x})\sqrt { \beta_1}}$, $C_1=\frac {8} {\log(1+\sigma^{-2)}}$, $\theta_x=\frac{\sum_{i=1}^{n}\theta_i}{n}$ and $\gamma_T$ It depends on the type of surrogate because the predicted uncertainty can differ depending on a model's ability to handle noisy measurements.
\end{theorem}

This indicates that by specifying $\delta$ and a total budget $\theta_T$, \tool \ can be configured to achieve a hypervolume error $\eta$ with confidence $1-\delta$.

\begin{proof}
Initially, using Lemma \ref{lem:vol_change} and \ref{lem:cauchy}, we show how the change of volume of the Pareto region $\sum_{t=1}^{T} \Delta V_{t,i}$ is related to the total budget $\theta_T$:

\begin{equation*}
    \begin{aligned}
    \sum_{t=1}^{T} \Delta {V}_{t,i} \leq 2\Big(\frac{C_1 \beta_T \gamma_T}{2}\Big)^{n/2}{\frac{(\frac{\theta_T}{\theta_x}(n+1))^{1/2}}{n!}} 
    \end{aligned}
\end{equation*}

Now, we relate hypervolume error $\eta$ and $\theta_T$. Let $1_n = (1, ... ,1)^T$ and let $e_i$ denote the $i^{th}$ canonical base vector. \shortcite{zuluaga2013active} that considers $a_i$ to be the maximum value for each $f_i(x)$, with probability $1 - \delta$.
Here, $a_i=\max_{\bm{x} \in \mathfrak{X_m}, 1 \leq i \leq n} {\sigma_i (\bm{x})\sqrt { \beta_1}}$ obtained from the width of the confidence regions, as shown in Figure \ref{fig:lem7}. We obtain this by replacing the co-variance term $k_i(\bm{x},\bm{x})$ used for measuring the width of the confidence region only for GP in Lemma 12 by ~\cite{zuluaga2013active} with variance $\sigma_i^2 (\bm{x})$ to extend our proof for both GP and RF surrogate models.
\textsc{PAL} \shortcite{zuluaga2013active} also showed that the projection $a_i$, where $1 \leq i \leq n$, onto the hyperplane $H_n$ is an n-simplex $S_n$ has a volume of $\frac {\sqrt{n}a^{n-1}} {(n-1)!}$. Hypervolume error $\eta$ depends on the distance between the boundaries defined by $\mathfrak{F}_{pess}$ and $\mathfrak{F}_{opt}$ at any iteration that is bounded by $\frac {\sqrt{n}a^{n-1}} {(n-1)!}{V}_t$~\cite{zuluaga2013active}.
At any iteration $t$, $V_t$ can be written as the difference between initial volume of the Pareto region $V_1$ and sum of change of volume $\Delta V_t$. At iteration $T$, hypervolume error $\eta$ can be written as the following:

\begin{equation*}
    \begin{aligned}
   \eta  & \leq  \frac {\sqrt{n}a^{n-1}} {(n-1)!} \left\{ V_1 - \sum_{t=1}^{T} \Delta V_{t,i} \right\} \\ 
   & \leq \frac {\sqrt{n}a^{n-1}} {(n-1)!} \left\{ a^n - 2\Big(\frac{C_1 \beta_T \gamma_T}{2}\Big)^{n/2}{\frac{(\frac{\theta_T}{\theta_x}(n+1))^{1/2}}{n!}}  \right\} \ \text{where} \ V_1=a^n\\
    \end{aligned}
\end{equation*}



\end{proof}

\begin{figure}[h]
    \centering
    \includegraphics[width=\textwidth]{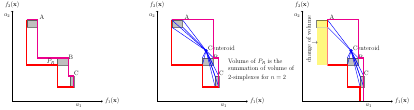}
    \caption{Example Pareto region $P_R$ for $n=2$ objectives (left). The Pareto region is a sum of n-simplexes (middle). The change of volume after evaluation across objective $f_1$ is shown by the yellow region (right).}
    \label{fig:centroid}
\end{figure}

\begin{lemma}
\label{lem:vol_change}
Given $\delta \in (0,1)$ and $ \beta_t= 2/9 \log(n|\mathfrak{X_m}|\pi^2t^2/{6\delta})$, the following holds with probability $\leq 1-\delta$ with $C_1=\frac{8}{\log(1+\sigma^{-2})}$.

\begin{equation}
    \sum_{t=1}^{T} \Delta {V}_{t,i}{^2} \leq \frac{4(C_1 \beta_T \gamma_T)^n}{(n!)^2}\frac{n+1}{2^n},
\end{equation}
\end{lemma}

\begin{proof} The change of volume of the Pareto region at any iteration $t$
    would be across only one objective $f_i$, where $1 \leq i \leq n$.
    Therefore, we need to determine the change of volume $\Delta V_{t,i}$ for
    $f_i$ only. Let us assume that $m$ is the centroid of the Pareto region $P_R$. If we add each vertex of the uncertainty region $R_t(\bm{x})$ of each design $\bm{x}$ with centroid $m$ an n-simplex is formed. So, the Pareto region $P_R$ can be shown as the sum of all these n-simplexes similar to Figure \ref{fig:centroid} (middle). When a design $\bm{x}_t$ is evaluated across an objective, volume of the Pareto region is reduced by the volume of two n-simplexes as shown by the yellow region in Figure \ref{fig:centroid} (right) i.e., an n-simplex becomes n-1-simplex due to reduction of uncertainty. Volume of an n-simplex is given by $\frac{s^n}{n!}\sqrt{\frac{n+1}{2^n}}$, where $s$ is the length of the side. 
\begin{equation*}
\begin{aligned}
  \Delta {V}_{t,i} & \leq  2V(\Delta^n)\\
  & \leq \frac{2s^n}{n!}\sqrt{\frac{n+1}{2^n}}\\
\end{aligned}
\end{equation*}
By using the width of the uncertainty region $2 \beta_t^{1/2} \sigma_{t-1,i}(\bm{x}_t)$ across an objective $f_i$ as the side length $s$ we get the following: 
\begin{equation*}
\begin{aligned}
  \Delta {V}_{t,i} & \leq \frac{2(2 \beta_t^{1/2} \sigma_{t-1,i}(\bm{x}_t))^n}{n!}\sqrt{\frac{n+1}{2^n}} \\
\end{aligned}
\end{equation*}
where $1 \leq i \leq n$. As $\beta_t$ is increasing the above equation can be written similar to \shortcite{zuluaga2013active} by the following:

\begin{equation*}
\begin{aligned}
  \Delta {V}_{t,i}^{2} & \leq \frac{4(4 \beta_T \sigma^2 ( \sigma^{-2}\sigma_{t-1,i}^2(\bm{x}_t)))^n}{(n!)^2} \frac{n+1}{2^n}\\
  & \leq \frac{4(4 \beta_T \sigma^2 C_2 \log(1+\sigma^{-2}\sigma_{t-1,i}^2(\bm{x}_t)))^n}{(n!)^2}\frac{n+1}{2^n},\\
\end{aligned}
\end{equation*}
where $C_2=\frac{\sigma^{-2}}{\log(1+\sigma^{-2})}$. Applying summation on the above we get  
\begin{equation*}
\begin{aligned}
  \sum_{t=1}^{T} \Delta {V}_{t,i}^{2}  \leq \frac{4(4 \beta_T \sigma^2 C_2 \sum_{t=1}^{T} \log(1+\sigma^{-2}\sigma_{t-1,i}^2(\bm{x}_t)))^n}{(n!)^2}\frac{n+1}{2^n}\\
\end{aligned}
\end{equation*}

With $C_1=8\sigma^2C_2$ we get the following:
\begin{equation*}
\begin{aligned}
  \sum_{t=1}^{T} \Delta V_{t,i}^{2}  & \leq  \frac{4(4\beta_T \sigma^2 C_2 \Delta V(\bm{y}_T;\bm{f}_T,i))^n}{(n!)^2}\frac{n+1}{2^n}\\
  & \leq \frac{4(C_1 \beta_T \Delta V(\bm{y}_T;\bm{f}_T))^n}{(n!)^2}\frac{n+1}{2^n} \\
  & \leq \frac{4(C_1 \beta_T \gamma_T)^n}{(n!)^2}\frac{n+1}{2^n}\\
\end{aligned}
\end{equation*}


\end{proof}
\begin{lemma}
\label{lem:cauchy}
Given $\delta \in (0,1)$ and $ \beta_t= 2/9 \log(n|\mathfrak{X_m}|\pi^2t^2/{6\delta})$, the following holds with probability $\leq 1-\delta$.

\begin{equation}
     \sum_{t=1}^{T} \Delta {V}_{t,i} \leq 2\Big(\frac{C_1 \beta_T \gamma_T}{2}\Big)^{n/2}{\frac{(\frac{\theta_T}{\theta_x}(n+1))^{1/2}}{n!}}   \ for \ T \geq 1
\end{equation}
\end{lemma}

\begin{proof} Similar to Lemma 6 in \shortcite{zuluaga2013active}, by applying Cauchy-Schwarz inequality on Lemma \ref{lem:vol_change} as $(\sum_{t=1}^{T}\Delta{V}_{t,i})^2 \leq T \sum_{t=1}^{T}\Delta {V}_{t,i}^2$, we obtain the following: 

\begin{equation}
    \sum_{t=1}^{T} \Delta {V}_{t,i} \leq 2\Big(\frac{C_1 \beta_T \gamma_T}{2}\Big)^{n/2}{\frac{(T(n+1))^{1/2}}{n!}} \ for  \ T \geq 1 
\label{eq:16}
\end{equation}

In the worst case, $T \leq \frac{\theta_T}{\theta_x}$, where  $\theta_x=\frac{\sum_{i=1}^{n}\theta_i}{n}$.




\end{proof}

\subsection{Runtime Complexity of \textsc{FlexiBO}}

We analyze the run-time complexity of \tool \ using Gaussian Processes (GP) and random forests (RF) as surrogate models, separately, in this section. The total complexity of \tool \ can be determined by combining complexities of \tool \ from modeling, Pareto region construction, and  sampling stages. 

Let $|\mathfrak{X}| = q$ be the total number of designs in the design space. However, we only consider $|\mathfrak{X}_m| = m$ designs sampled by Monte-Carlo sampling in this approach. We also consider $N_0 + t$ designs to train the surrogate models at each iteration $t$. Let us consider $s = m + N_0 + t$. Note that by design, $s << q$. As a result, our \tool \ algorithm is significantly faster.

\noindent \paragraph{Modeling.} In the modeling stage, only a small subset of designs are used to train the surrogate models at each iteration. Training a GP with $s$ number of designs takes $\mathcal{O}(s^3+ms^2)$ \shortcite{rasmussen2003gaussian} time. Training a RF with $s$ designs takes $\mathcal{O}(n_{t}n_vs^2 \log s)$ time where $n_{t}$ is the number of trees, and $n_v$ is the number of features used at each level. Determining the uncertainty region of each design $\bm{x} \in \mathfrak{X_m}$ takes an additional $\mathcal{O}(m)$ time. Therefore, total complexity of the modeling stage for using GP surrogate model is  $\mathcal{O}(s^3+ms^2+ m)$ and for RF surrogate model is $\mathcal{O}(n_{t}n_vs^2 \log s + m)$. 

\noindent \paragraph{Pareto Region Construction.}
In the Pareto region construction stage, we initially determine the non-dominated designs in the design space and later use the non-dominated designs $\bm{x} \in \mathfrak{U}$ to construct the Pareto fronts. The complexity of finding the non-dominated designs is $\mathcal{O}(m^2)$. The complexity of constructing the Pareto fronts is similar to the complexity of determining the number of designs on the boundary of the convex hull, which can be performed in $\mathcal{O}(m\log m)|$ time if the number of objectives $n=2$. When $n > 2$,
the Pareto fronts are constructed in $\mathcal{O}(m (\log m)^{n-2}+
m\log m)$ time \shortcite{kung1975finding}.   

\noindent \paragraph{Sampling.}
In the sampling stage, \tool \ determines the next sample $\bm{x}_{t}$ and objective $f_{t,i}$ for evaluation. To do so, \tool \ computes the acquisition function $\alpha_{t,i}(\bm{x})$ for each design in $\mathfrak{X_m}^*$ and selects the maximum.
To compute $\alpha_{t,i}(\bm{x})$ across an objective $f_i$,
\tool \ needs to compute the volume of the Pareto region $P_R$ by updating the
uncertainty values of $\bm{x}$ in $\mathfrak{X_m}^*$. This would take $\mathcal{O}(m)$ time as  $|\mathfrak{X_m}|^* = m$ in the worst case.
After measuring the selected design $\bm{x}_{t}$ across objective $f_{t,i}$, we update the evaluated designs set $S_i$ and objective evaluation cost $\theta_{t,i}$. All of these are done in constant time and we can safely ignore them in our analysis. Therefore, the total run-time complexity of \tool \ in the sampling stage is $O(m)$, regardless of the surrogate model.


\noindent \paragraph{Overall.}
Finally, we determine the overall complexity of \tool \ using GP and RF for $n$ objectives by combining the complexities of the three stages discussed above. 
When GP is used as the surrogate model, the total complexity of \tool \ for $n=2$ objectives is $\mathcal{O}(s^3+ms^2+m^2+m \log m+2m)$ and for $n \geq 3$ objectives the total complexity is $\mathcal{O}(s^3+ms^2+m^2+m(\log m)^{n-2} + m \log m + 2m)$. To simplify these expressions, we consider $s=m$. Now, the total complexity for $n=2$ objectives using GP surrogate model is approximately $\mathcal{O}(s^3+s^2+s \log s+s)$ and for $n=3$ objectives is $\mathcal{O}(s^3+s^2+s(\log s)^{n-2} + s \log s + s)$.

Similarly when RF is used as a surrogate model, the total complexity of \tool \ for $n=2$ objectives is $ \mathcal{O}(n_t n_v s^2 \log s+m^2+m \log m+2m)$ and for $n \geq 3$ objectives is $ \mathcal{O}(n_t n_v s^2 \log s+m^2+m(\log m)^{n-2} + m \log m + 2m)$. After simplification the complexity for $n=2$ objectives can be written as  $\mathcal{O}(s^2 \log s+s^2+s \log s+s)$ and for $n=3$ objectives the complexity can be rewritten as $\mathcal{O}(s^2 \log s+s^2+s(\log s)^{n-2} + s \log s + s)$. 

Upon further simplification, we observe that the complexity of \tool \ with the GP surrogate model is approximately $\mathcal{O}(s^3)$ and for the RF surrogate model the complexity is $\mathcal{O}(s^2 \log s)$, where $s$ is significantly lower than the total number of designs $q$. 

\end{document}